\RequirePackage{fix-cm}

\documentclass[twocolumn]{svjour3}          
\smartqed  
\usepackage{graphicx}
\usepackage{amsmath,amssymb,amsfonts}
\usepackage{graphicx}
\usepackage{textcomp}
\usepackage{adjustbox}
\usepackage{multirow}
\usepackage{makecell, longtable}
\usepackage[square,numbers]{natbib}
\usepackage{bbold} 
\usepackage[table,xcdraw]{xcolor} 
\definecolor{Gray}{gray}{0.85}
\usepackage{hhline}
\usepackage{breqn} 
\usepackage{floatrow} 
\usepackage[figurename=Figure]{caption} 
\usepackage{subcaption} 
\usepackage{algorithmic}
\usepackage{algorithm}

\usepackage{booktabs} 
\usepackage{makecell} 

\usepackage{siunitx}  
\usepackage{url}
\usepackage[hidelinks]{hyperref} 
\hypersetup{
	colorlinks   = true, 
	urlcolor     = blue, 
	linkcolor    = blue, 
	citecolor   = blue 
} 

\def\makeheadbox{\relax}
\begin{document}

\title{Multivariate Anomaly Detection based on Prediction Intervals Constructed using Deep Learning
}

\titlerunning{Anomaly Detection Using Deep Learning}        

\author{Thabang Mathonsi$^1$         \and
        Terence L. van Zyl$^2$ 
}

\authorrunning{T. Mathonsi, T.L. van Zyl} 

\institute{$^1$T. Mathonsi (corresponding author)\at
	School of Computer Science and Applied Mathematics \\
\emph{University of the Witwatersrand}\\
Johannesburg, South Africa \\
              \email{thabang.mathonsi@wits.ac.za}           
           \and
           $^2$T.L. van Zyl \at
              Institute for Intelligent Systems\\
          \emph{University of Johannesburg}\\
          Johannesburg, South Africa \\
      \email{tvanzyl@uj.ac.za}}

\date{}

\maketitle

\begin{abstract}
It has been shown that deep learning models can under certain circumstances outperform traditional statistical methods at forecasting. Furthermore, various techniques have been developed for quantifying the forecast uncertainty (prediction intervals). In this paper, we utilize prediction intervals constructed with the aid of artificial neural networks to detect anomalies in the multivariate setting. Challenges with existing deep learning-based anomaly detection approaches include $(i)$ large sets of parameters that may be computationally intensive to tune, $(ii)$ returning too many false positives rendering the techniques impractical for use, $(iii)$ requiring labeled datasets for training which are often not prevalent in real life. Our approach overcomes these challenges. We benchmark our approach against the oft-preferred well-established statistical models. We focus on three deep learning architectures, namely, cascaded neural networks, reservoir computing and long short-term memory recurrent neural networks. Our finding is deep learning outperforms (or at the very least is competitive to) the latter.
\end{abstract}

\section{Introduction}
\citet{Hawk80} defines an outlier as an observation that "{\it deviates so significantly from other observations as to arouse suspicion that a different mechanism generated it.}"
Anomalies, also known as outliers, arise for several reasons, such as intentional acts of malice, collapse or failure of various systems, or fraudulent activity. Detecting these anomalies in a timeous manner is essential in various decision-making processes. Anomaly detection is not to be confused with novelty detection, which is focused on identifying new patterns in the data. However, the techniques used for anomaly detection are often used for novelty detection and vice versa. \citet{Car20} draw a comparison between the two fields and detail recent advances in the methods applicable to each field.

Anomaly detection, when framed as a classification problem, has been modelled using One-Class Support Vector Machines as far back as the works of \citet{Schol99}. Dealing with a highly imbalanced class distribution (where normal observations far exceed anomalous data points), \citeauthor{Schol99} detect novel patterns by looking at the local density distribution of normal data.

\citet{Mal15} use stacked LSTM networks for anomaly detection in time series. They train the network on non-anomalous data and use it as a predictor over several time steps.  The resulting prediction errors are modelled as a multivariate Gaussian distribution, which is then used to assess the likelihood of anomalous behaviour. Later, \citet{Mal16} examine anomaly detection for mechanical devices in the presence of exogenous effects. An LSTM-based Encoder-Decoder scheme is employed that learns to reconstruct time-series behaviour under normalcy and after that uses reconstruction error to detect anomalies. \citet{Sch17} employ a deep convolutional generative adversarial network in an unsupervised manner to learn a manifold of normal anatomical variability and apply it to the task of anomaly detection to aid disease detection and treatment monitoring.

More recently, the research has focused on time-series anomalies that manifest over a period of time rather than at single time points. These are known as range-based or collective anomalies. \citet{Nguyen18} consider these and use prediction errors from multiple time-steps for detecting collective anomalies.

Others, such as \citet{Zhu17} use Bayesian Neural Networks (BNNs, \cite{Khera18}) and the Monte Carlo (MC) dropout framework to construct prediction intervals, inspired by \citet{Gal16a} as well as \citet{Gal16b}.

Several probability-based methods have also been successfully applied to the detection of anomalies \cite{Agg13}. If said statistical methods have been applied successfully, why then consider deep learning for anomaly detection at all? We provide three reasons. Firstly, the boundary between normal and anomalous behaviour is at times not precisely defined, or it evolves continually, i.e. the distribution of the data changes dynamically as new instances are introduced to the time series. In such instances, statistical algorithms may sub-perform \cite{Uth17} unless, for instance, they are retrained continuously to mitigate the changes in the data distributions. Secondly, deep learning algorithms have been shown to scale well in other applications \cite{Bri16}, and we hypothesise that this scalability attribute may extend to cases that require large-scale anomaly detection.  Finally, in the most recent M4 competition, it is shown that deep learning produces better prediction intervals than statistical methods~\cite{Makri20a}. This superior prediction interval result is corroborated by the works of~\citet{Smyl20} in the univariate case as well as~\citet{Mat1} in the multivariate setting, for instance. Our hypothesis is that better prediction intervals can be used to build better anomaly detection machinery.

\subsection{Related Work}

There have been immense advances made in applying deep learning methods in fields such as forecasting~\cite{Makri20a, Makri20b}, but there is a relative scarcity of deep learning approaches for anomaly detection and its sub-disciplines~\cite{Kwon19, Thu20}.

Considering anomaly detection as a supervised learning problem requires a dataset where each data instance is labeled. With the anomaly detection problem phrased as such, there are general steps usually followed in the literature. Although not identical in their methodologies, even the most recent advances are not dissimilar to approaches first adopted by~\citet{Mal15},~\citet{Chau15} and~\citet{Mal16}, for instance. Generally, methodologies incorporate the following steps:
\begin{enumerate}
	\item use a deep learning algorithm or architecture (e.g. LSTM, encoder-decoder) for generating forecasts, retain the errors as they are used later for anomaly detection;
	\item assume the resultant errors follow a Gaussian (or other) distribution;
	\item learn a threshold by maximizing either the $F$-score, precision, recall or some combination thereof on the validation set.
\end{enumerate}

There are also some comprehensive reviews, including, for instance, the works of~\citet{Ade16}, who survey deep learning-based methods for fraud detection,~\citet{Kwon19} who focus on techniques and applications related to cyber-intrusion detection, and as recent as~\citet{Thu20} who focus on the challenge of detecting anomalies in the domain of big data.

All of these reviews, and the research highlighted above, touch on some critical points related to drawbacks.  We summarize these drawbacks in Section~\ref{motivate} below. 

To avoid ambiguity, all the techniques described in this Section shall further be referred to as "reviewed techniques."

\subsection{Motivation}
\label{motivate}

All the reviewed techniques, as well as their classical statistical counterparts, have similar drawbacks.  Classical techniques such as statistical and probabilistic models are typically suitable only for univariate datasets \cite{Makri20a, Smyl20}. For the more realistic multivariate case, \citet{Li16} for instance, propose a method for analyzing multivariate datasets by first transforming the high-dimensional data to a uni-dimensional space using Fuzzy C-Means clustering. They then detect anomalies with a Hidden Markov Model. However, \citeauthor{Li16}'s approach does not fully utilize the additional information provided by the data's correlation.

Another issue with the reviewed techniques is their reliance on datasets that contain the ground truth, where the anomalies have been labeled.  These labels are sometimes also referred to as golden labels. There are difficulties associated with gathering this kind of data in real life, such as the cost of labeling or human error in the labeling process. Even once the reviewed models are trained and parameterized on such data, if available, the dynamic nature of most real-life scenarios would mean the tuned review model might not generalize to new types of anomalies unless retraining is performed continuously \cite{Uth17}.

\subsection{Contribution}
\begin{itemize}
	\item We introduce a dynamic threshold-setting approach, which learns and adapts the applicable threshold as new information becomes available.
	\item No golden labels are required for training or detecting the thresholds, and subsequently detecting the anomalies.
	\item We consolidate two streams of research by testing our assertions on both outlier and step-change anomalies, whereas previous researchers mostly focused on either one of these.
	\item We critically analyze forecast machinery, construct competitive prediction intervals using computational means where no statistical methodology is readily applicable, then use all of this information to detect anomalies.
\end{itemize}

\section{Datasets}
Our datasets come from a diverse array of economic sectors. We use data from various categories ranging from climate to space aviation. The most important aspect when selecting our datasets is ensuring they are distinctly different in terms of size, attributes, and the composition of normal to anomalous ratio of observations. This data sourcing methodology allows us to forecast for varied horizons, and mitigates biases in model performance. All the datasets are also freely available and have been used extensively in research related to ours and other fields.

\subsection{Air Quality Dataset.} We use the Beijing air quality dataset (AQD, \cite{uci14}) from UCI. It contains measurements of hourly weather and pollution level for five years spanning January 1$^{\text{st}}$, 2010 to December 31$^{\text{st}}$, 2014 at the US embassy in Beijing, China.  Meteorological data from the Beijing Capital International Airport are also included.  The attribute of interest is the level of pollution in the air (pm2.5 concentration), while Table \ref{data_1} details the complete feature list.

\begin{table}[h!tbp]
	\centering
	\caption{Air Quality Data Description.}
	\label{data_1}
	\begin{tabular}{r|l}
		\toprule
		\thead{Attribute} &  \thead{Description}   \\
        \bottomrule\toprule
		pm2.5        &   pm2.5 concentration   \\ 
		DEWP        &        dew point        \\ 
		TEMP        &       temperature       \\ 
		PRES        &        pressure         \\ 
		cbwd        & combined wind direction \\ 
		Iws         &  cumulated wind speed   \\ 
		Is         & cumulated hours of snow \\ 
		Ir         & cumulated hours of rain \\ 
		\bottomrule
	\end{tabular} 
\end{table}

\subsection{Power Consumption Dataset}
The second dataset is UCI's Household Power Consumption (HPC, \cite{uci12}), a multivariate time series dataset detailing a single household's per-minute electricity consumption spanning December 2006 to November 2010.

With the date and time fields collapsed into a unique identifier, the remaining attributes are described in Table \ref{data_02}.

\begin{table}
	\centering
	\caption{Power Consumption Data Description.}
	\label{data_02}
	\resizebox{\columnwidth}{!}{%
	\begin{tabular}{ r|l}
		\toprule
		\thead{Attribute}      & \thead{Description}                                    \\ 
		\bottomrule\toprule
		global\_active\_power   & total active power consumed (\si{\kilo\watt})                 \\ 
		global\_reactive\_power & total reactive power consumed (\si{\kilo\watt})               \\ 
		global\_intensity       &  \makecell[lc]{minute-averaged current intensity (\si{\ampere})} \\ 
		sub\_metering\_1        & active energy for kitchen (\si{\watt\hour}) \\ 
		sub\_metering\_2        & active energy for laundry (\si{\watt\hour})  \\ 
		sub\_metering\_3        & \makecell[lc]{active energy for climate control systems  (\si{\watt\hour})}	\\ 
		\bottomrule
	\end{tabular} 
	}
\end{table}

A new attribute is created:
\begin{dmath}
	\text{sub\_metering\_4} = \dfrac{100}{6}\times \text{global\_active\_power} - \sum_{k = 1}^{3}\text{sub\_metering\_} k.
\end{dmath} It represents the per-minute active energy consumption not already captured by sub\_meterings 1, 2 and 3. 

\subsection{Bike Sharing Dataset}

Next, we consider the UCI Bike Sharing Dataset (BSD, \cite{uci13}).  It catalogues for each hour the number of bikes rented by customers of a bike-sharing service in Washington DC for the 2011 and 2012 calendar years.  It includes features such as whether a particular day is a national holiday and the weather and seasonal information. Besides the date-time and index attributes that we collapse into a unique identifier, the dataset has 17 attributes in total.  We select only a subset for our analysis, as described in Table \ref{data_03}.

\begin{table}[h!tbp]
	\caption{Bike Sharing Data Description.}
	\label{data_03}
	\centering
	\resizebox{\columnwidth}{!}{%
	\begin{tabular}{r|l}
		\toprule
		\thead{Attribute} &  \thead{Description}  \\
		\bottomrule\toprule
		holiday        &   boolean whether day is holiday or not   \\ 
		weekday        &        day of the week        \\  
		weathersit     &  \makecell[lc]{
		                        1 if clear, few clouds, partly cloudy                  \\
		                        2 if cloudy, mist, broken clouds, few clouds           \\
			                    3 if light snow or rain, thunderstorm, scattered clouds\\
			                    4 if heavy rain, ice pallets, thunderstorm, snow mist, fog
		                   } \\   
		hum        &       humidity       \\  
		temp        &        normalized temperature in degrees Celsius 
		\\  
		casual        & count of casual users \\  
		registered         &  count of registered users   \\  
		cnt         & count of total rental bikes, sum of casual and registered \\
		\bottomrule
	\end{tabular} 
	}
\end{table}

\subsection{Bearings Dataset}
We also use NASA's Bearings Dataset (BDS, \cite{nasa1}).  It contains sensor readings on multiple bearings that are operated until failure under a constant load causing degradation.  These are one second vibration signal snapshots recorded at ten minute intervals per file, where each file contains $20,480$ sensor data points per bearing.  It contains golden labels and has been considered extensively by other researchers in their experimentation; see, for example, \citet{Pro17}.

\subsection{Webscope Benchmark Dataset}
\label{sub:WBD}
Lastly, we also evaluate our approach using the Yahoo Webscope Benchmark dataset (YWB, \cite{yahoo1}), which is considered a good data source in the research community by, for instance, \citet{Thi17}. It comprises large datasets with golden labels.  It is helpful for our purposes as it contains anomalies based on outliers and change-points. The first benchmark, A1, is real site traffic to Yahoo and its related products.  The other three, A2, A3, and A4, are synthetic. They all contain three months worth of data-points though of varying lengths, capture frequencies, seasonality patterns and the total number of labeled anomalies.

Because we synthesize anomalies for AQD, BSD and HPC, we only use A1 from YWB. Using the other synthetic benchmarks in YWB may not add further insights to our analysis.

\section{Methodology}
\label{meth}

Our approach utilizes three fundamental steps. We:
\begin{enumerate}
	\item forecast using deep learning and probability-based models,
	\item derive prediction intervals using computational means and built-in density distributions, respectively, and
	\item use said prediction intervals to detect anomalies.
\end{enumerate}

Each of the fundamental steps, and intermediate ones, are outlined below.

\subsection{Forecasting}

\subsubsection{AQD}
We frame a supervised learning problem where, given the weather conditions and pollution measurements for the current and $23$ prior hours, $t, t - 1, t - 2, \hdots,  t -23$, we forecast the pollution at the next hour $t + 1$.

\subsubsection{HPC}
\begin{figure}[h!tbp]
	\centering 
	\resizebox{\columnwidth}{!}{%
	\includegraphics[width=0.99\columnwidth]{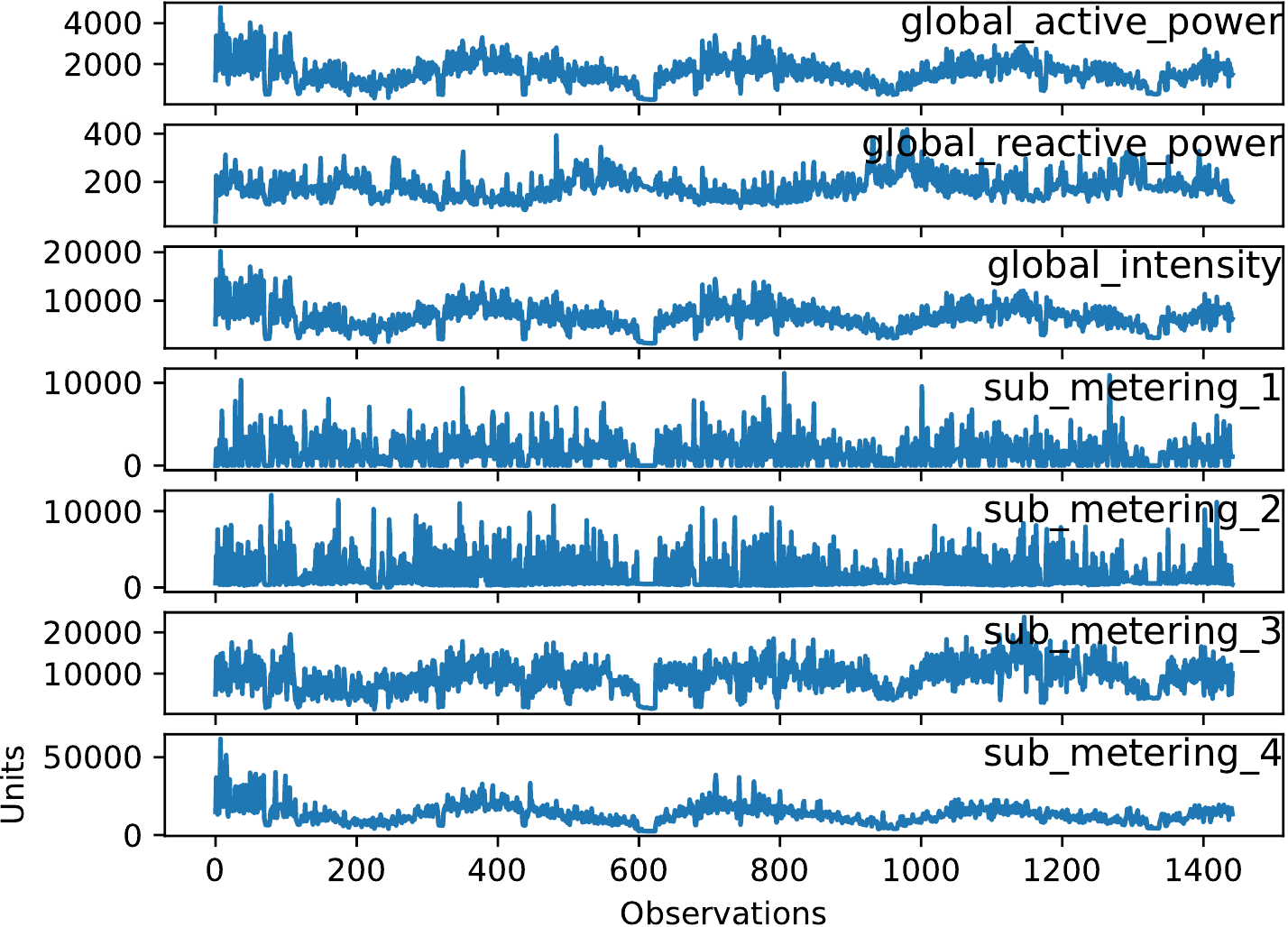}
	}
	\caption{Separate Attribute Plots of the Power Consumption Data Spanning Four Years.}
	\label{fig:refineddata_02} 
\end{figure}

Given power consumption for 4 weeks, we forecast it for the week ahead. The observations are converted from per-minute to daily aggregates, as illustrated in Figure~\ref{fig:refineddata_02}. Seven values comprise a single forecast, one value for each day in the standard week (beginning on a Sunday and ending Saturday). We evaluate each forecast in this aggregated setting as well as with each time step separately.

There are only 205 samples available for training, and this may pose a problem for parameters-tuning. So we disregard the standard week, and use the prior $28$ days to forecast the coming $7$ days.  This flattening and use of overlapping windows effectively transforms the $205$ samples into a total of $1,418$.

In contrast, the testing framework remains unchanged, i.e. we use the prior four standard weeks to forecast the daily power consumption for the subsequent standard week.

\subsubsection{BSD}
Given all available information for BSD for the past 3 months, we forecast the total rental bikes for the coming week.

\subsubsection{BDS}
Given the sensor readings for the past 10 hours, we forecast the readings for the next hour. The mechanical degradation in the bearings occurs gradually over time, so we conduct our analysis using only data points occurring every 10 minutes instead of per-second observations. In total, we have $20,480$ observations across each of the four bearings.

To easily detect the point where the bearing vibration readings begin oscillating dramatically, we transform the signal from the time- to the frequency domain using a Fourier transform.

\begin{figure}[h!t]
	\centering 
	\includegraphics[width=1\columnwidth]{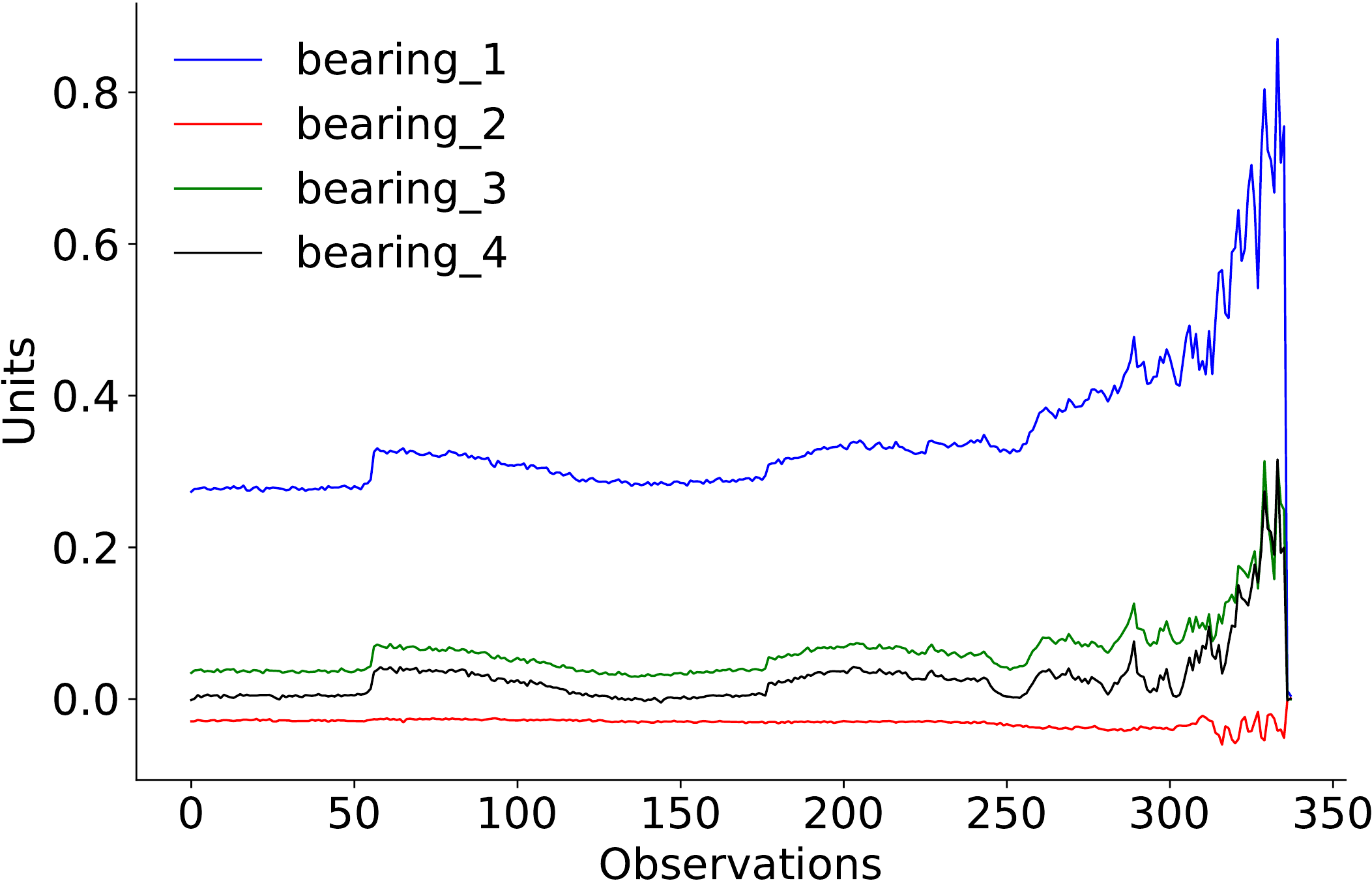}\caption[Bearing Sensor Test Frequency Data]{Bearing Sensor Test Frequency Data.} 
	\label{fig:bearings1} 
\end{figure}

As depicted in Figure~\ref{fig:bearings1}, there is an increase in the frequency amplitude for readings progressing to the complete degradation and subsequent failure of each bearing.

\subsubsection{YWB}
Lastly, for the YWB, we utilize the last five days worth of observations to forecast traffic to the Yahoo entities for the next hour.

Note, all the above forecast problems can be reframed using different window sizes.  We state our chosen window sizes here for reproducibility. These windows were chosen using empirical studies with optimizing based on training accuracy.

For each dataset, we use the relevant forecast framework as described above and subsequently construct prediction intervals with all three deep learning algorithms and all three statistical algorithms.

To a great extent, the treatment of each dataset is not dissimilar. As such, the narrative that follows might overlap.  For ease of exposition, all methods are the same unless explicitly stated.

\subsection{Preprocessing}
We undertake the following steps in preparing the time series data for forecasting. Here we only highlight the most used techniques for all the datasets. NA values are removed. NA is distinct from missing (blank) values, the latter amounting to $4.664\%$ for AQD and $1.252\%$ for HPC, for instance, are imputed using the iterative imputer from the Scikit-learn module \cite{sci11}, which models each feature with missing values as a function of other features in a round-robin fashion.  We set the imputer to parameterize only on past values for each iteration, thus avoiding information leakage from the future testing set.

We label encode categorical values. Label encoding is chosen over one-hot encoding for instances where there are many categories in total (more than five), as the former reduces memory consumption. Furthermore, we normalize the datasets.

All the datasets are composed of time series with different seasonality patterns.  In such instances, ANNs tend to struggle with seasonality \citep{Clav15, Smyl20}. A standard remedy is to deseasonalize the series first and remove the trend using a statistical procedure like the Seasonal and Trend decomposition using Loess (STL, \cite{Cleve90}). For a more detailed treatment of the pre-processing stage, see \citet{Mat1}.

\subsection{Algorithms}
\label{algo}

For benchmarking, the statistical methods used are Multiple Linear Regression (MLR, \cite{Mat2005}), Vector Autoregression Moving-Average with Exogenous Regressors (VARMAX, \cite{Han88}) and Seasonal Autoregressive Integrated Moving-Average with Exogenous Regressors (SARIMAX, \cite{Aru16}).

The deep learning architectures used are the Cascaded Neural Networks based on cascade-correlation (CCN), first published by~\citet{Fahlman90}, a variant of the Recurrent Neural Network (RNN, \cite{Will89}) known as long short-term memory (LSTM, \cite{Hoch97}), and one class of reservoir computing, Echo State Networks (ESN, \cite{Jaeger01}).

For aiding in the construction of prediction intervals, we employ Monte Carlo Dropout \cite{Gal16b} and the quantile bootstrap \cite{Dav13} also known as the reverse percentile bootstrap \cite{Hest14}.  Details on implementation follow in Section~\ref{sub:drop} and \ref{sub:boot} respectively.

\subsection{Measures of Accuracy}
\label{sub:MoA}

For the point forecasts we use the Root Mean Squared Error (RMSE) that gives error in the same units as the forecast variable itself.

\begin{align}
\text{RMSE} &= \sqrt{\dfrac{1}{n}\sum_{t = 1}^{n}\left( y_t - \hat{y}_t \right)^2},
\end{align} where there are $n$ forecasts, $y_t$ is the out-of-sample observation and $\hat{y}_t$ the forecast at time $t$.

For the prediction intervals we use the Mean Interval Score (MIS, \cite{Gneit07}), averaged over all out-of-sample observations. 

\begin{dmath}
\label{eq:MIS}
\text{MIS$_{\alpha}$} = \dfrac{1}{n}\sum_{t = 1}^{n}\left( U_t - L_t \right) + \dfrac{2}{\alpha} \left( L_t - y_t \right) \mathbb{1}_{\left\lbrace y_t < L_t \right\rbrace} + \dfrac{2}{\alpha} \left( y_t - U_t \right) \mathbb{1}_{\left\lbrace y_t > U_t \right\rbrace},
\end{dmath} 
where $U_t$ $\left( L_t \right)$ is the upper (lower) bound of the prediction interval at time $t$, $\alpha$ is the significance level and $\mathbb{1}$ is the indicator function.

The MIS$_{\alpha}$ adds a penalty at the points where future observations are outside the specified bounds. The width of the interval at $t$ is added to the penalty, if any. As such, it also penalizes wide intervals.  Finally, this sum at the individual points is averaged.

As a supplementary metric, we employ the Coverage Score (CS$_{\alpha}$) which indicates the percentage of observations that fall within the prediction interval,
\begin{align}
\text{CS$_{\alpha}$} &= \dfrac{1}{n}\sum_{t = 1}^{n} \mathbb{1}_{\left\lbrace U_t \leq y_t \leq L_t \right\rbrace}.
\end{align}
The subscript $\alpha$ denotes explicit dependence on the significance level.

For evaluating the performance when it comes to anomaly detection we employ precision, recall and the $F_1$-score. We also use the $Ed$-score~\cite{Buda17}, which scores the early detection capability of an algorithm in the range $0$ (worst) to $1$ (best). An increase (decrease) of say $5\%$ in the $Ed$-score measure may be interpreted as a technique detecting an anomaly on average $5\%$ of the time interval earlier (later). The $Ed$-score for an algorithm $X_i$ for anomaly $a$ that falls within window $W$ for dataset $D$ is given by

\begin{equation}
Ed\text{-score}\left(X_i \right) = 
\begin{cases}
\dfrac{e - \text{idx}\left(X_i\left(a\right)\right)}{e - b} &\;\text{if }a \in A^{X_i}_D\\
0 &\;\text{if }a \in A^{X_i}_D\\
\end{cases}
\end{equation}

where $A^{X_i}_D$ denotes the set of anomalies detected by algorithm $X_i$ in dataset $D$, $X_i\left(a\right)$ is the time taken for the initial detection of anomaly $a$ in $D$ by algorithm $X_i$, and idx$\left(X_i\left(a\right)\right)$ is the index of that timestamp, $b$ and $e$ are the beginning and end indices for window $W$, respectively. We use the indices of the timestamp instead of the time difference as the datasets use different time steps between observations. 

\subsection{Analysis}
\label{subsub:analysis}

\subsubsection{AQD}

We configure the LSTM with $50$ neurons in the first and only hidden recurrent layer, and $1$ neuron in the output layer for predicting pollution.  The model is trained for $25$ epochs with a batch size of $72$. The input shape is $24$ time steps with $8$ features.

The ESN is initialized with $384$ input units, $1$ output unit, and $4,096$ internal network units in the dynamic reservoir ($\mathcal{O} \left( 10 \text{ x } \text{\# input units}\right) $).  We use a $24$ hour wash-out interval.

The CCN is initialized with $72$ neurons in the input layer and one neuron in the output layer.  The cascading is capped at $24$ layers.

\begin{figure*}[h!tbp]
	\centering 
	\resizebox{\columnwidth}{!}{%
	\includegraphics[width=1\textwidth]{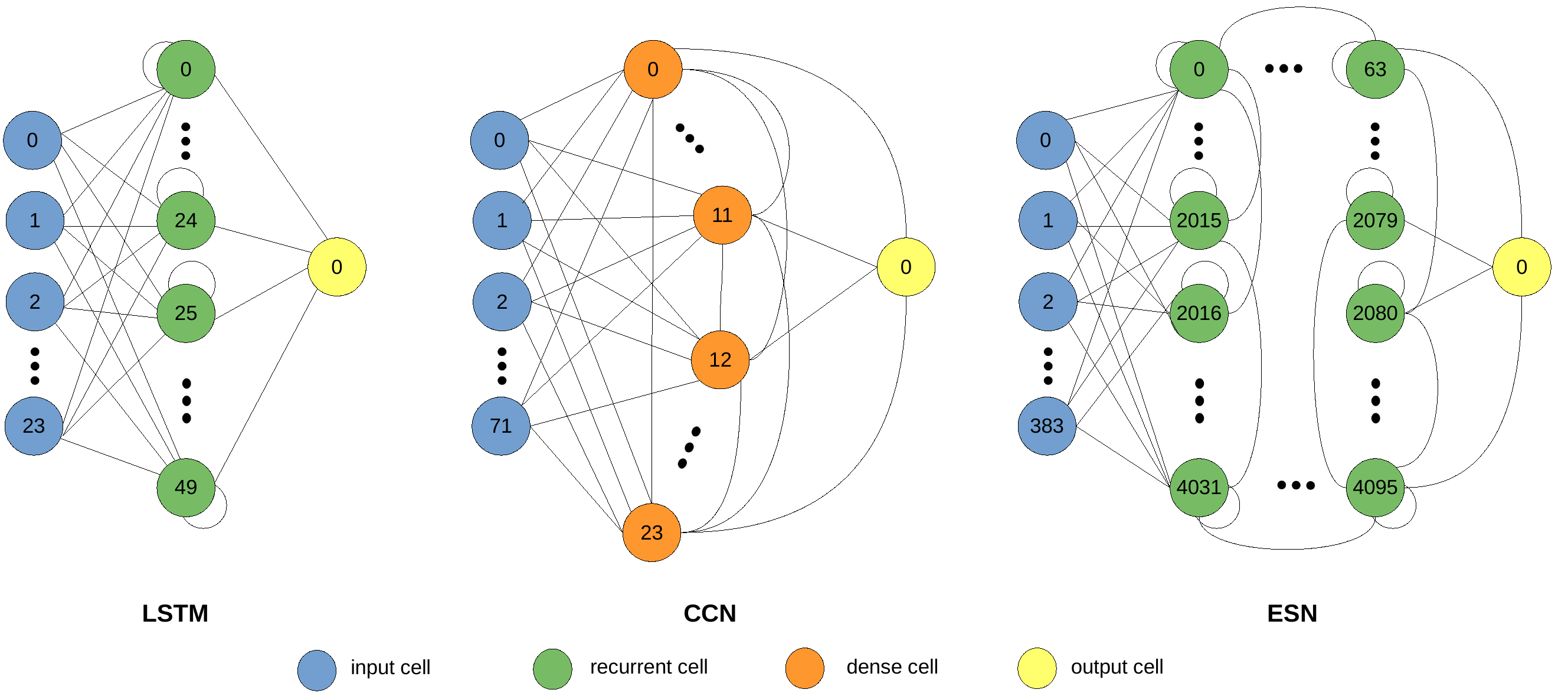}}\caption{Flowchart Diagrams Depicting our Deep Learning Model Architecture Configurations for AQD.}
	\label{fig:arch} 
\end{figure*}

The structure of the deep learning architectures for AQD are given in Figure~\ref{fig:arch} as an illustrative example. The CCN starts off with one hidden cell that is fully connected. As additional cells are added (if required), they are connected to the input and output nodes, as well as each of the preceding hidden nodes. The ESN's reservoir is composed of recurrent cells that are sparsely connected. The deep learning architecture configurations for the other datasets are detailed below.

\subsubsection{HPC}

Our model is developed with one hidden LSTM recurrent layer with 200 neurons. The second layer is fully connected with 200 neurons.  Lastly, the direct forecast is produced by an output layer that outputs a vector with seven entries, as previously described. Our model is trained for 65 epochs with a batch size of 18, determined experimentally to be optimal.

The ESN uses a $28$ day wash-out interval.  The CCN has seven neurons in the output layer, with cascading capped at $28$ layers.

The first three calendar years are used as training data and the final year for model evaluation purposes.

\subsubsection{BSD}
The LSTM is initialized with 100 neurons in the sole hidden recurrent layer, followed by an output layer that directly forecasts a vector with seven days worth of forecasts.  We fit the model for $25$ epochs with a batch size of $32$.

The ESN uses a seven day wash-out interval.  The CCN has $168$ neurons in the output layer, with cascading capped at $28$ layers.

\subsubsection{BDS}
We instantiate the LSTM with 100 neurons in the input layer, one hidden recurrent layer with 50 neurons, and an output layer comprised of 4 neurons.  Our model is fit for 100 epochs with a batch size of 32.

The ESN is initialized with $48$ input units, $4$ output units, and $1,024$ internal network units in the dynamic reservoir. We use a $24$ hour wash-out interval.

The CCN is composed of $100$ neurons in the input layer and $4$ neurons in the output layer. The cascading is capped at $24$ layers.

\subsubsection{YWB}
The LSTM has 5 hidden layers with 4, 16, 48, 16, and 4 neurons, respectively. Our model is fit for 45 epochs with a batch size of 24.

The ESN uses a $3$ day wash-out interval. The CCN has $4$ neurons in the output layer, with cascading capped at $9$ layers.

\subsubsection{Overall}

In terms of AQD, we use holdout cross-validation, fitting each of the models on the first four years of data and testing with the remaining year.  For all the other datasets, we use walk-forward validation for optimizing the hyper-parameters. For instance, with HPC, our models forecast a week ahead $\hat{y}_{t + 1}$.  After that, the actual observations from a week ahead $y_{t + 1}$ are presented to the models, and are used in the set of observations $ \hdots y_{t - 2}, y_{t - 1}, y_{t}, y_{t + 1}$ for the current iteration in the walk-forward validation process, before we forecast the subsequent week $\hat{y}_{t + 2}$. 

As an additional method to mitigate computational cost, the models are evaluated statically, i.e. they are trained on the historical observations and then iteratively forecast each step of the walk-forward validation.

For each architecture, we use the Mean Absolute Error (MAE, \cite{Sam10}) loss function and the efficient Adam version of stochastic gradient descent \cite{King14}. Each configuration was selected over various others based on empirical studies of test RMSE reduction and configuration behaviour over time.

The input variables to VARMAX and SARIMAX are declared as exogenous.

After each model is fit, we forecast, replace the seasonality and trend components, invert the scaling, and compute an error score for each. The results follow in Section~\ref{sub:forecast}.

\subsection{Evaluation}	

The LSTM is implemented using the LSTM and Dense class of the Keras API v2.1.5 \citep{Cho15} for Python 3.7 using TensorFlow v1.15.0 framework as backend \citep{Abadi16}.  The code implementation for the CCN is written from scratch using Python 3.7, and ESN is implemented using PyTorch v0.3 \citep{PAs17} for Python 3.7. The na{\"i}ve or benchmark methods are implemented using the Statsmodels API \cite{sea10} for Python 3.7. All models are trained on an Intel Core i7-9750H processor with an Nvidia GeForce GTX 1650 GPU.

\subsection{Recurrent Layer Dropout}
\label{sub:drop}

\begin{figure}[h!t]
	\centering 
	\includegraphics[width=0.9\columnwidth]{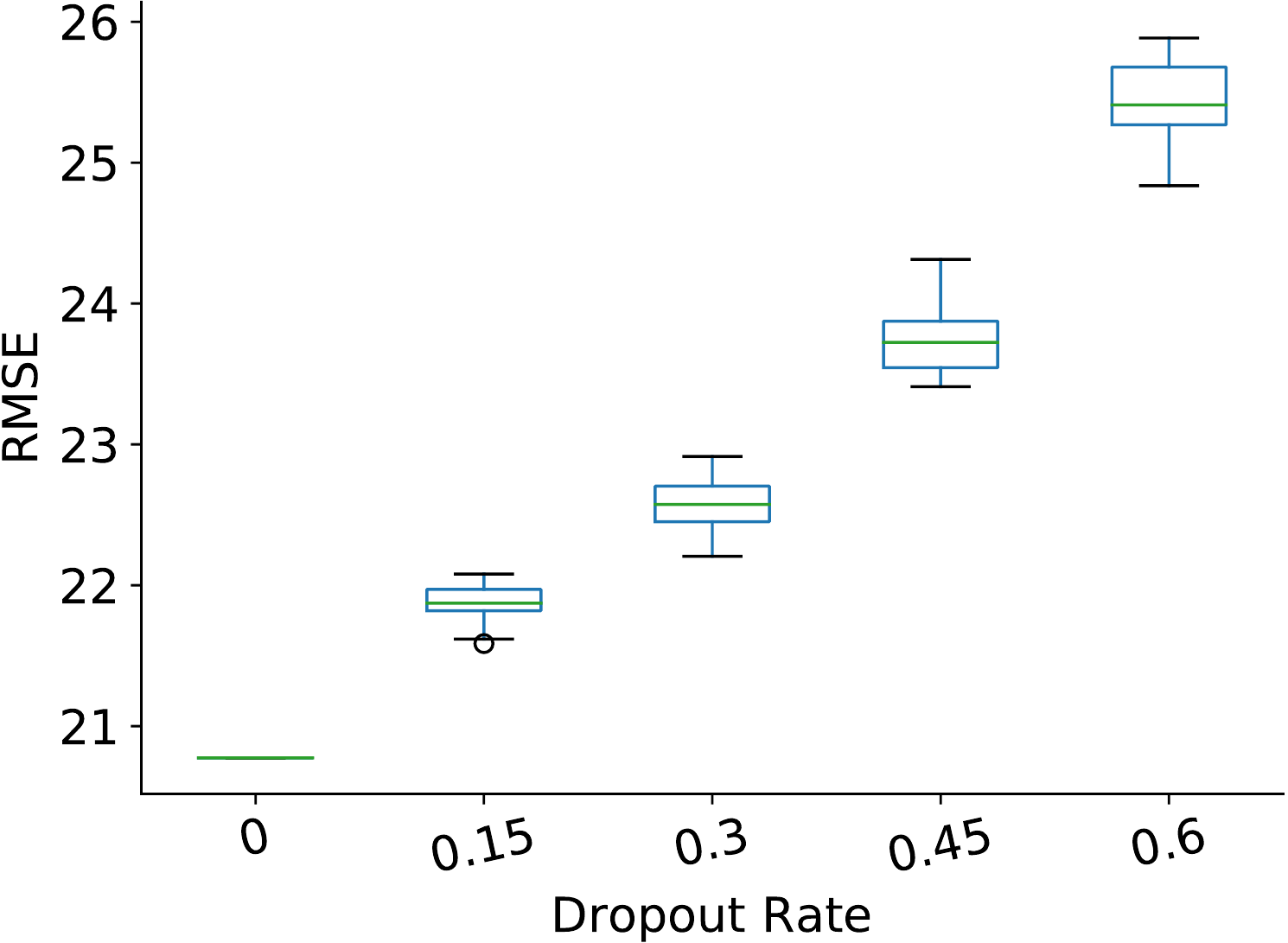}\caption[Dropout results distributions]{Dropout RMSE Score Distributions.} 
	\label{fig:dropoutdistribution} 
\end{figure}

We compare the forecast accuracy at dropout rates of $15\%$, $30\%$, $45\%$, and $60\%$ to the standard LSTM fit above (i.e. with no dropout). Each dropout experiment is repeated $35$ times to mitigate the stochastic nature of the deep learning algorithms. Table~\ref{tab:dropout_01}, and the box and whisker plot in Figure~\ref{fig:dropoutdistribution}, display the distributions of results for each configuration in the case of AQD as an illustrative example.

\begin{table}[h!t]
	\centering
	\caption{Dropout RMSE Score Distributions.}
	\label{tab:dropout_01}
	\begin{tabular}{lrrrrr}
		\toprule
		& \multicolumn{5}{c}{\thead{Dropout Rate}} \\ [\aboverulesep]
		&
		{\thead{0.0}} &
		{\thead{0.15}} &
		{\thead{0.3}} &
		{\thead{0.45}} &
		{\thead{0.6}} \\ 
		\bottomrule \toprule
		\multicolumn{1}{l|}{{\textbf{count}}} &
		{35} &
		{35} &
		{35} &
		{35} &
		{35} \\ 
		\multicolumn{1}{l|}{{\textbf{mean}}} &
		{20.7746} &
		{21.8681} &
		{22.5086} &
		{23.7820} &
		{25.4886} \\ 
		\multicolumn{1}{l|}{{\textbf{std}}} &
		{0.0000} &
		{0.1283} &
		{0.1793} &
		{0.2436} &
		{0.2812} \\ 
		\multicolumn{1}{l|}{{\textbf{min}}} &
		{20.7746} &
		{21.5665} &
		{22.2587} &
		{23.4037} &
		{24.8913} \\ 
		\multicolumn{1}{l|}{{\textbf{25\%}}} &
		{20.7746} &
		{21.8193} &
		{22.4969} &
		{23.5335} &
		{25.2295} \\ 
		\multicolumn{1}{l|}{{\textbf{50\%}}} &
		{20.7746} &
		{21.8543} &
		{22.5990} &
		{23.7627} &
		{25.4671} \\ 
		\multicolumn{1}{l|}{{\textbf{75\%}}} &
		{20.7746} &
		{21.9773} &
		{22.7260} &
		{23.8939} &
		{25.6549} \\ 
		\multicolumn{1}{l|}{{\textbf{max}}} &
		{20.7746} &
		{22.0783} &
		{22.9699} &
		{24.3575} &
		{25.8718} \\ \bottomrule
	\end{tabular}
\end{table}

From Table~\ref{tab:dropout_01}, we note that the standard implementation (LSTM with no dropout) outperforms all the configurations with dropout. At $15\%$ the output has the tightest distribution (a desirable property for constructing prediction intervals). The $15\%$ dropout configuration has a competitive RMSE compared to the standard implementation.  Because we drop out at such a low probability, this configuration retains a lot more information than the others. We thus use this configuration for prediction interval construction.

We run the chosen configuration $100,000$ times. In this case, the $15\%$ dropout probability. We forecast, and use the distribution of said forecast values to construct our prediction intervals. To compute the $(1 - \alpha) \times 100 \%$ prediction interval, we use the percentiles at 
\begin{dmath*}
\left(\dfrac{\alpha}{2}, \left(1 - \alpha\right) + \left( \dfrac{\alpha}{2}\right) \right)
\end{dmath*}.

\subsection{Bootstrap}
\label{sub:boot}

We resample residuals in the following manner:

\begin{enumerate}
	\item Once the model is fit, retain the values fitted, $y^{\prime}_t$, and residuals $\varepsilon_t = y_t - y^{\prime}_t, t = 1,\dots, m$, where $m \leq n$ our forecast horizon.
	\item For each tuple, $\left(\vec{x}_t,y_t \right)$, in which $\vec{x}_t$ is the multivariate explanatory variable, add the randomly resampled residual, $\varepsilon_j$, to the fitted value $y^{\prime}_t$, thus creating synthetic response variables $y^*_t = y^{\prime}_t + \varepsilon_j$ where $j$ is selected randomly from the list $\left(1, ..., m\right)$ for every $t$.
	\item Using the values $\left(\vec{x}_t,y^*_t \right)$, forecast $\hat{y}$, then refit the model using synthetic response variables $y^*_t$, again retaining the residuals and forecasts.
	\item Repeat steps 2 and 3, $100,000$ times.
\end{enumerate}

We use the distribution of the forecast values to construct the prediction intervals using percentiles in the same manner described in Section~\ref{sub:drop} above. As in the case of the recurrent layer dropout, this technique is advantageous as it retains the explanatory variables' information.

\subsection{Anomaly Detection}

Anomalies are labeled in YWB. BDS does not contain any golden labels. We synthesize anomalies by superimposing them on the other three datasets, as previously mentioned in Section~\ref{sub:WBD} above. This step is undertaken in a stochastic manner, where a typical composition is given in Table~\ref{tab:data_01}.

\begin{table}[h!t]
	\centering
	\caption{Anomaly Data Structure Summary.}
	\label{tab:data_01}
	\resizebox{1.00\columnwidth}{!}{%
	\begin{tabular}{l|clc}
		\toprule
		\thead{Data} & \thead{\# Metrics} & \thead[l]{Type} & \thead{Contamination} \\ 
		\bottomrule\toprule
		\textbf{AQD}  & \tablenum{10}             & changepoint                & \tablenum{0.0133}  \\ 
		\textbf{HPC}  & \tablenum{7}              & collective                 & \tablenum{0.0142}  \\ 
		\textbf{BSD}  & \tablenum{17}             & contextual                 & \tablenum{0.0209}  \\ 
		\textbf{BDS}  & \tablenum{36}             & changepoint                & \tablenum{0.0191}  \\ 
		\textbf{YWB}  & \tablenum{67}             & \makecell[l]{contextual \& collective}   & \tablenum{0.0176}  \\ \bottomrule
	\end{tabular}
	}
\end{table}

If a new observation falls within our constructed prediction interval, it is classified as normal.  If it falls outside the interval, we:
\begin{enumerate}
	\item determine by which magnitude it falls outside the bounds. This measure is deducible using a variant of MIS defined earlier (Equation~\ref{eq:MIS} in Section~\ref{sub:MoA}), i.e. the Interval Score (IS)
	    \begin{align*}
	    \qquad \text{IS}_\alpha &= \left( U_t - L_t \right) + \dfrac{2}{\alpha} \left( L_t - y_t \right) \mathbb{1}_{\left\lbrace y_t < L_t \right\rbrace}\\
	    &\qquad~+ \dfrac{2}{\alpha} \left( y_t - U_t \right) \mathbb{1}_{\left\lbrace y_t > U_t \right\rbrace}
	    \end{align*}
	    If this measure is at least $33\%$ larger than that of the previous observation that fell outside bounds, then we
	\item employ Chebyshev’s inequality \cite{Dix68} which guarantees a useful property for a wide class of probability distributions, i.e. no more than a certain fraction of values can be more than a certain distance from the mean. A suitable threshold can be tuned to avoid many false alerts. In our case, we utilize this inequality to identify that $99\%$ of the values must lie within ten times the standard deviation.
\end{enumerate}
If an out-of-sample observation falls outside the prediction interval bounds and falls beyond ten times the standard deviation of the observations (our dynamic threshold, $\leq 1\%$ of all observations), it is classified as anomalous.

\begin{figure*}[h!tbp]
	\centering 
	\includegraphics[width=0.8\textwidth]{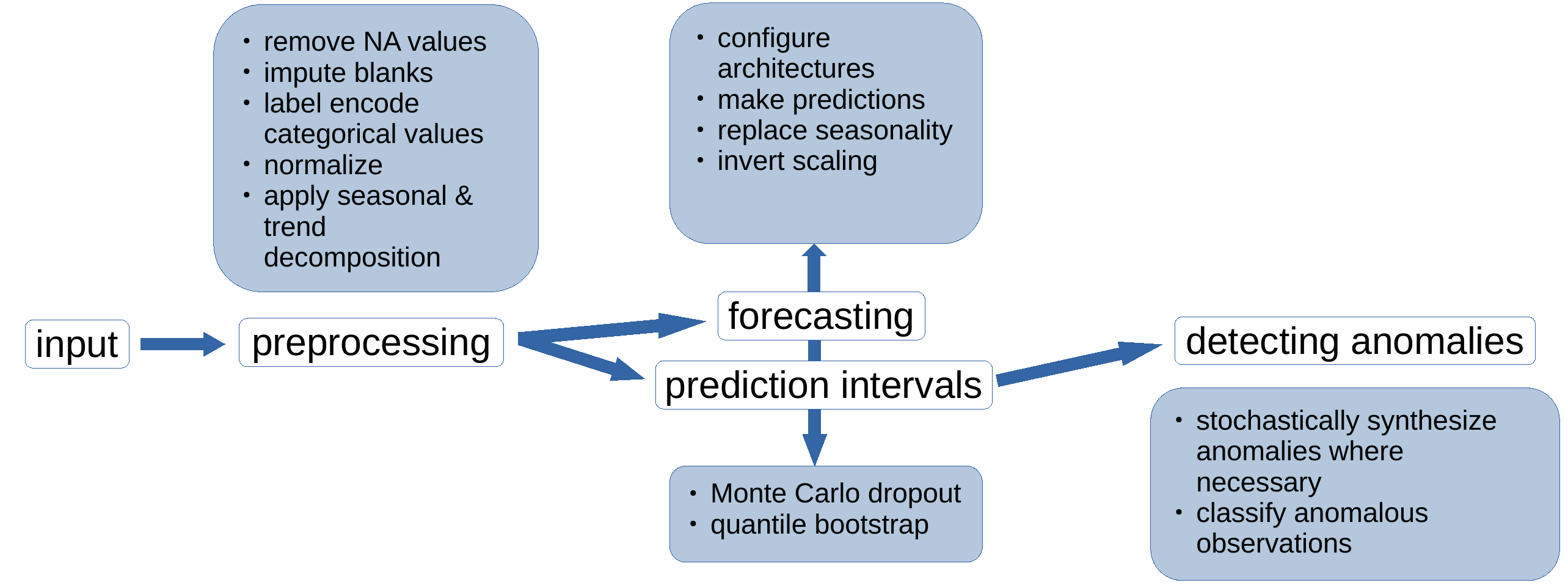}\caption{Flowchart Diagram Depicting our Deep Learning Model Architectures.} 
	\label{fig:dataflow} 
\end{figure*}

Essentially, the anomaly detection procedure may be summarized in Algorithm~\ref{algo:ad}, and the entire structure of our deep learning models and data flow process is presented graphically in Figure~\ref{fig:dataflow}.

\begin{algorithm}
\caption{Anomaly Detection}\label{algo:ad}

\begin{algorithmic}
\IF{$U_t \leq y_t \leq L_t$}
    \STATE $y_t$ is normal
\ELSE
    \IF{$\text{IS}_\alpha \left( y_t \right) \geq 1.33 \times \text{IS}_\alpha \left( y_* \right)$ \AND $ y_t > 10 \times \text{std} {\left\lbrace \dots, y_{t-3}, y_{t-2}, y_{t-1} \right\rbrace}$}
    \STATE $y_t$ is anomalous
    \COMMENT{where $y_*$ is the last anomalous observation}
    \ENDIF
\ENDIF
\end{algorithmic}
\end{algorithm}

\section{Results}

\subsection{Forecasting}\label{sub:forecast}

A persistence model~\cite{Coi13} gives an RMSE score of $30$, $465$, $94$, $1.8$, and $301$ respectively for AQD, HPC, BSD, BDS and YWB. Those from the models under consideration, truncated to four decimals, are shown in Table~\ref{tab:rmse}.

\begin{table}[h!t]
	\caption{Model Error Scores.}
	\label{tab:rmse}
	\centering
	\begin{adjustbox}{width=\columnwidth}
		\begin{tabular}{ll|rrrrr}
			\toprule
			\multicolumn{2}{l}{} & \multicolumn{5}{c}{\thead{Data}}                                    \\ [\aboverulesep]
			\multicolumn{2}{l}{} & \thead{AQD} & \thead{HPC} & \thead{BSD} & \thead{BDS} & \thead{YWB} \\ 
			\bottomrule\toprule
			\multirow{3}{*}{\textbf{\makecell[l]{Statistical\\Benchmarks}}} & MLR     & 24.8791      & 415.8602     & 43.8581      & 1.5001       & 278.6595     \\ 	                                                                & SARIMAX & 22.9377      & 377.8571     & 34.3472      & 1.3712       & 251.8902     \\                                                                 & VARMAX  & 23.1093      & 401.3319     & 39.2279      & 1.4879       & 266.0144     \\ \midrule\midrule
			\multirow{3}{*}{\textbf{\makecell[l]{Deep\\ Learning}}}          & LSTM    & 20.1342      & 337.9452     & 31.8951      & 0.9588       & 213.8532     \\ 	                                                                 & ESN     & 21.2357      & 349.1773     & 31.0451      & 1.0057       & 209.3517     \\                                                                  & CCN     & 22.9876      & 392.5459     & 35.6843      & 1.1421       & 227.1096     \\ \bottomrule
		\end{tabular}
	\end{adjustbox}
\end{table}

\subsection{Prediction Intervals}

For ease of interpretation, we modify the MIS to introduce a scaling factor, such that the best performing model has scaled Mean Interval Score (sMIS) equal to $1$, and the rest are larger than $1$. The worst performing model thus has the largest sMIS value. For each significance level $\alpha$ considered we have,
\begin{align}
\text{sMIS}_{\alpha, i} &= \dfrac{\text{MIS}_{\alpha, i}}{\min\limits_{i \in X} \text{MIS}_{\alpha, i}},
\end{align}
where $X$ comprises all algorithms considered (both statistical and deep learning). We further highlight the top two performers at each significance level.

\begin{table}[h!t]
	\centering
	\caption{Prediction Interval Performance for Air Quality Data.}
	\label{tab:results_02}
	\begin{adjustbox}{width=\columnwidth}
		\begin{tabular}{ll|rrr|rrr}
			\toprule
			\multicolumn{2}{c}{} &
			\multicolumn{3}{c|}{\thead{sMIS $\alpha$}} &
			\multicolumn{3}{c }{\thead{CS $\alpha$}} \\ [\aboverulesep]
			\multicolumn{2}{l|}{} &
			\multicolumn{1}{c|}{\textbf{0.1}} &
			\multicolumn{1}{c|}{\textbf{0.05}} &
			\multicolumn{1}{c|}{\textbf{0.01}} &
			\multicolumn{1}{c|}{\textbf{0.1}} &
			\multicolumn{1}{c|}{\textbf{0.05}} &
			\multicolumn{1}{c }{\textbf{0.01}} \\ 
			\bottomrule\toprule
			\multicolumn{1}{l}{} &
			MLR &
			{1.1704} &
			{1.2105} &
			{1.2323} &
			{0.9031} &
			{0.9171} &
			{0.9305} \\
			\multicolumn{1}{l}{} &
			SARIMAX &
			\cellcolor{Gray}1.0778 &
			{1.2093} &
			{1.2167} &
			\cellcolor{Gray} 0.9524 &
			\cellcolor{Gray} 0.9668 &
			\cellcolor{Gray} 0.9797 \\ 
			\multicolumn{1}{l}{\multirow{-3}{*}{\begin{tabular}[c]{@{}l@{}}\textbf{Statistical}\\ \textbf{Bench-}\\ \textbf{marks}\end{tabular}}} &
			VARMAX &
			{1.4141} &
			{1.2662} &
			{1.0881} &
			{0.9449} &
			{0.9621} &
			{0.9761}\\ 
			\midrule\midrule
			\multicolumn{1}{l}{} &
			LSTM &
			\cellcolor{Gray} 1.0000 &
			{1.1437} &
			\cellcolor{Gray} 1.0000 &
			\cellcolor{Gray} 0.9620 &
			\cellcolor{Gray} 0.9810 &
			\cellcolor{Gray} 0.9824 \\ 
			\multicolumn{1}{l}{} &
			ESN &
			{1.1859} &
			\cellcolor{Gray} 1.1365 &
			{1.1367} &
			{0.9179} &
			{0.9308} &
			{0.9435} \\ 
			\multicolumn{1}{l}{\multirow{-3}{*}{\begin{tabular}[c]{@{}l@{}}\textbf{Deep}\\ \textbf{Learning}\end{tabular}}} &
			CCN &
			{1.2103} &
			\cellcolor{Gray} 1.0000 &
			\cellcolor{Gray} 1.0655 &
			{0.9094} &
			{0.9285} &
			{0.9633} \\ 
			\bottomrule
		\end{tabular}
	\end{adjustbox}
\end{table}


\begin{table}[h!t]
	\centering
	\caption{Prediction Interval Performance for Bike Sharing Data.}
	\label{tab:results_03}
	\begin{adjustbox}{width=\columnwidth}
		\begin{tabular}{ll|rrr|rrr}
			\toprule
			\multicolumn{2}{c}{} &
			\multicolumn{3}{c|}{\thead{sMIS $\alpha$}} &
			\multicolumn{3}{c }{\thead{CS $\alpha$}} \\ [\aboverulesep]
			\multicolumn{2}{l|}{} &
			\multicolumn{1}{c|}{\textbf{0.1}} &
			\multicolumn{1}{c|}{\textbf{0.05}} &
			\multicolumn{1}{c|}{\textbf{0.01}} &
			\multicolumn{1}{c|}{\textbf{0.1}} &
			\multicolumn{1}{c|}{\textbf{0.05}} &
			\multicolumn{1}{c }{\textbf{0.01}} \\ 
			\bottomrule\toprule
			\multicolumn{1}{l}{} &
			MLR &
			{1.2131} &
			{1.1688} &
			{1.2072} &
			{0.9062} &
			{0.9169} &
			{0.9165} \\ 			\multicolumn{1}{l}{} &
			SARIMAX &
			{1.0364} &
			{1.2052} &
			{1.2708} &
			{\cellcolor{Gray}0.9493} &
			{0.9501} &
			{\cellcolor{Gray}0.9553} \\ 
			\multicolumn{1}{l}{\multirow{-3}{*}{\begin{tabular}[c]{@{}l@{}}\textbf{Statistical}\\ \textbf{Bench-}\\ \textbf{marks}\end{tabular}}} &
			VARMAX &
			{1.1236} &
			{1.1796} &
			{1.1676} &
			{0.9075} &
			{0.9107} &
			{0.9194} \\ 
			\midrule \midrule
			\multicolumn{1}{l}{} &
			LSTM &
			{1.1727} &
			{1.0872} &
			{1.1662} &
			{0.9370} &
			{\cellcolor{Gray}0.9519} &
			{0.9422} \\ 
			\multicolumn{1}{l}{} &
			ESN &
			{\cellcolor{Gray}1.0000} &
			{\cellcolor{Gray}1.0000} &
			{\cellcolor{Gray}1.0000} &
			{\cellcolor{Gray}0.9587} &
			{\cellcolor{Gray}0.9683} &
			{\cellcolor{Gray}0.9772} \\
			\multicolumn{1}{l}{\multirow{-3}{*}{\begin{tabular}[c]{@{}l@{}}\textbf{Deep}\\ \textbf{Learning}\end{tabular}}} &
			CCN &
			{\cellcolor{Gray}1.0186} &
			{\cellcolor{Gray}1.0388} &
			{\cellcolor{Gray}1.0409} &
			{0.9468} &
			{0.9472} &
			{0.9495} \\ \bottomrule 
		\end{tabular}
	\end{adjustbox}
\end{table}


\begin{table}[h!t]
	\centering
	\caption{Prediction Interval Performance for Household Power Consumption Data.}
	\label{tab:results_04}
	\begin{adjustbox}{width=\columnwidth}
		\begin{tabular}{ll|rrr|rrr}
			\toprule
			\multicolumn{2}{c}{} &
			\multicolumn{3}{c|}{\thead{sMIS $\alpha$}} &
			\multicolumn{3}{c }{\thead{CS $\alpha$}} \\ [\aboverulesep]
			\multicolumn{2}{l|}{} &
			\multicolumn{1}{c|}{\textbf{0.1}} &
			\multicolumn{1}{c|}{\textbf{0.05}} &
			\multicolumn{1}{c|}{\textbf{0.01}} &
			\multicolumn{1}{c|}{\textbf{0.1}} &
			\multicolumn{1}{c|}{\textbf{0.05}} &
			\multicolumn{1}{c }{\textbf{0.01}} \\ 
			\bottomrule\toprule
			\multicolumn{1}{l}{} &
			MLR &
			{1.2262} &
			{1.2123} &
			{1.1686} &
			{0.9081} &
			{0.9117} &
			{0.9144} \\ 
			\multicolumn{1}{l}{} &
			SARIMAX &
			{1.1081} &
			{\cellcolor{Gray}1.0181} &
			{1.1012} &
			{\cellcolor{Gray}0.9413} &
			{0.9514} &
			{0.9760} \\ 
			\multicolumn{1}{l}{\multirow{-3}{*}{\begin{tabular}[c]{@{}l@{}}\textbf{Statistical}\\ \textbf{Bench-}\\ \textbf{marks}\end{tabular}}} &
			VARMAX &
			{1.1764} &
			{1.1565} &
			{1.1148} &
			{0.9178} &
			{0.9262} &
			{0.9322} \\ 
			\midrule \midrule
			\multicolumn{1}{l}{} &
			LSTM &
			{\cellcolor{Gray}1.0449} &
			{1.1005} &
			{\cellcolor{Gray}1.0211} &
			{\cellcolor{Gray}0.9507} &
			{\cellcolor{Gray}0.9594} &
			{\cellcolor{Gray}0.9767} \\ 
			\multicolumn{1}{l}{} &
			ESN &
			1.0910 &
			{\cellcolor{Gray}1.0000} &
			{\cellcolor{Gray}1.0000} &
			{0.9372} &
			{\cellcolor{Gray}0.9516} &
			{\cellcolor{Gray}0.9895} \\ 
			\multicolumn{1}{l}{\multirow{-3}{*}{\begin{tabular}[c]{@{}l@{}}\textbf{Deep}\\ \textbf{Learning}\end{tabular}}} &
			CCN &
			{\cellcolor{Gray}1.0000} &
			{1.1452} &
			{1.0708} &
			{0.9157} &
			{0.9299} &
			{0.9571} \\ \bottomrule 
		\end{tabular}
	\end{adjustbox}
\end{table}


\begin{table}[h!t]
	\centering
	\caption{Prediction Interval Performance for Bearings Data.}
	\label{tab:results_05}
	\begin{adjustbox}{width=\columnwidth}
		\begin{tabular}{ll|rrr|rrr}
			\toprule
			\multicolumn{2}{c}{} &
			\multicolumn{3}{c|}{\thead{sMIS $\alpha$}} &
			\multicolumn{3}{c }{\thead{CS $\alpha$}} \\ [\aboverulesep]
			\multicolumn{2}{l|}{} &
			\multicolumn{1}{c|}{\textbf{0.1}} &
			\multicolumn{1}{c|}{\textbf{0.05}} &
			\multicolumn{1}{c|}{\textbf{0.01}} &
			\multicolumn{1}{c|}{\textbf{0.1}} &
			\multicolumn{1}{c|}{\textbf{0.05}} &
			\multicolumn{1}{c }{\textbf{0.01}} \\ 
			\bottomrule\toprule
			\multicolumn{1}{l}{} &
			MLR &
			{1.3562} &
			{1.2020} &
			{1.1549} &
			{0.9106} &
			{0.9134} &
			{0.9287} \\ 
			\multicolumn{1}{l}{} &
			SARIMAX &
			{\cellcolor{Gray}1.0000} &
			{1.1254} &
			{1.1832} &
			{0.8957} &
			{0.9009} &
			{0.9071}  \\ 
			\multicolumn{1}{l}{\multirow{-3}{*}{\begin{tabular}[c]{@{}l@{}}\textbf{Statistical}\\ \textbf{Bench-}\\ \textbf{marks}\end{tabular}}} &
			VARMAX &
			{1.1764} &
			{1.1565} &
			{1.2149} &
			{0.9075} &
			{0.9161} &
			{0.9193} \\ 
			\midrule \midrule
			\multicolumn{1}{l}{} &
			LSTM &
			{\cellcolor{Gray}1.0541} &
			{1.1045} &
			{\cellcolor{Gray}1.0299} &
			{\cellcolor{Gray}0.9401} &
			{\cellcolor{Gray}0.9412} &
			{\cellcolor{Gray}0.9491} \\ 
			\multicolumn{1}{l}{} &
			ESN &
			{1.0910} &
			{\cellcolor{Gray}1.0000} &
			{\cellcolor{Gray}1.0000} &
			{0.9202} &
			{\cellcolor{Gray}0.9346} &
			{\cellcolor{Gray}0.9398} \\ 
			\multicolumn{1}{l}{\multirow{-3}{*}{\begin{tabular}[c]{@{}l@{}}\textbf{Deep}\\ \textbf{Learning}\end{tabular}}} &
			CCN & 
			{1.1081} &
			{\cellcolor{Gray}1.0181} &
			{1.1012} &
			{\cellcolor{Gray}0.9213} &
			{0.9277} &
			{0.9301}\\ 
			\bottomrule
		\end{tabular}
	\end{adjustbox}
\end{table}


\begin{table}[h!t]
	\centering
	\caption{Prediction Interval Performance for Webscope Benchmark Data.}
	\label{tab:results_06}
	\begin{adjustbox}{width=\columnwidth}
		\begin{tabular}{ll|rrr|rrr}
			\toprule
			\multicolumn{2}{c}{} &
			\multicolumn{3}{c|}{\thead{sMIS $\alpha$}} &
			\multicolumn{3}{c }{\thead{CS $\alpha$}} \\ [\aboverulesep]
			\multicolumn{2}{l|}{} &
			\multicolumn{1}{c|}{\textbf{0.1}} &
			\multicolumn{1}{c|}{\textbf{0.05}} &
			\multicolumn{1}{c|}{\textbf{0.01}} &
			\multicolumn{1}{c|}{\textbf{0.1}} &
			\multicolumn{1}{c|}{\textbf{0.05}} &
			\multicolumn{1}{c }{\textbf{0.01}} \\ 
			\bottomrule\toprule
			\multicolumn{1}{l}{} &
			MLR &
			{1.3054} &
			{1.112} &
			{1.1499} &
			{0.8878} &
			{0.8923} &
			{0.8971} 
			\\ 
			\multicolumn{1}{l}{} &
			SARIMAX &
			{1.1081} &
			{\cellcolor{Gray}1.0169} &
			{1.1012} &
			{\cellcolor{Gray}0.9293} &
			{0.9313} &
			{0.9360} \\ 
			\multicolumn{1}{l}{\multirow{-3}{*}{\begin{tabular}[c]{@{}l@{}}\textbf{Statistical}\\ \textbf{Bench-}\\ \textbf{marks}\end{tabular}}} &
			VARMAX &
			{1.2262} &
			{1.2123} &
			{1.1686} &
			{0.8981} &
			{0.9108} &
			{0.9142}\\ 
			\midrule \midrule
			\multicolumn{1}{l}{} &
			LSTM &
			{1.0812} &
			{\cellcolor{Gray}1.0000} &
			{\cellcolor{Gray}1.0000} &
			{0.9271} &
			{\cellcolor{Gray}0.9360} &
			{\cellcolor{Gray}0.9494} 
			\\ 
			\multicolumn{1}{l}{} &
			ESN &
			{\cellcolor{Gray}1.0546} &
			{1.1050} &
			{\cellcolor{Gray}1.0233} &
			{\cellcolor{Gray}0.9307} &
			{\cellcolor{Gray}0.9394} &
			{\cellcolor{Gray}0.9464}\\ 
			\multicolumn{1}{l}{\multirow{-3}{*}{\begin{tabular}[c]{@{}l@{}}\textbf{Deep}\\ \textbf{Learning}\end{tabular}}} &
			CCN &
			{\cellcolor{Gray}1.0000} &
			{1.1452} &
			{1.0708} &
			{0.9157} &
			{0.9299} &
			{0.9309} \\ \bottomrule 
		\end{tabular}
	\end{adjustbox}
\end{table}

\begin{figure}[h!tbp]
	\centering 
	\includegraphics[width=0.9\columnwidth]{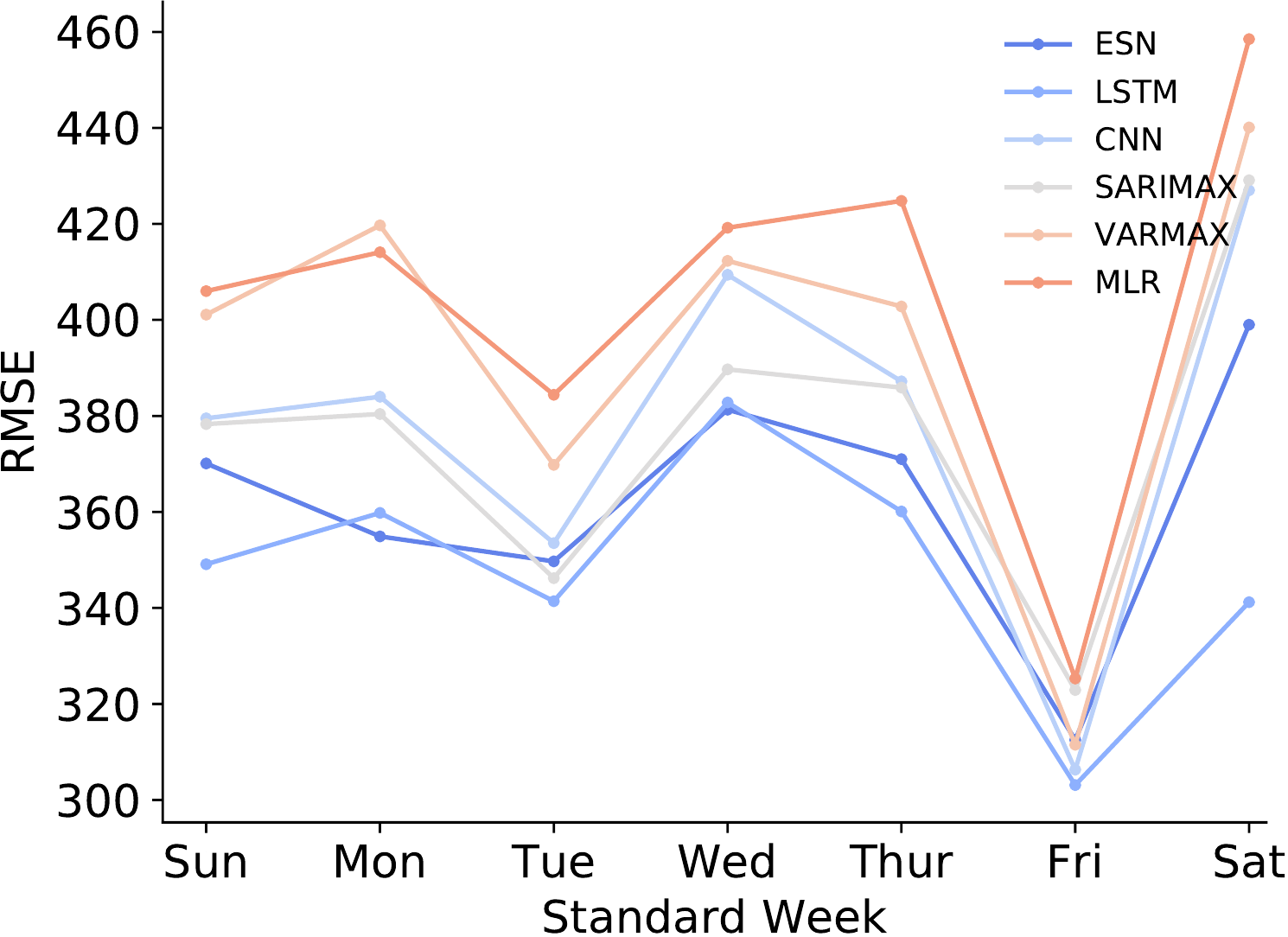}\caption[Daily RMSE]{Overall Daily RMSE for Household Power Consumption Dataset.} 
	\label{fig:dailyrmse} 
\end{figure}

\begin{figure}[h!tbp]
	\centering 
	\includegraphics[width=\textwidth]{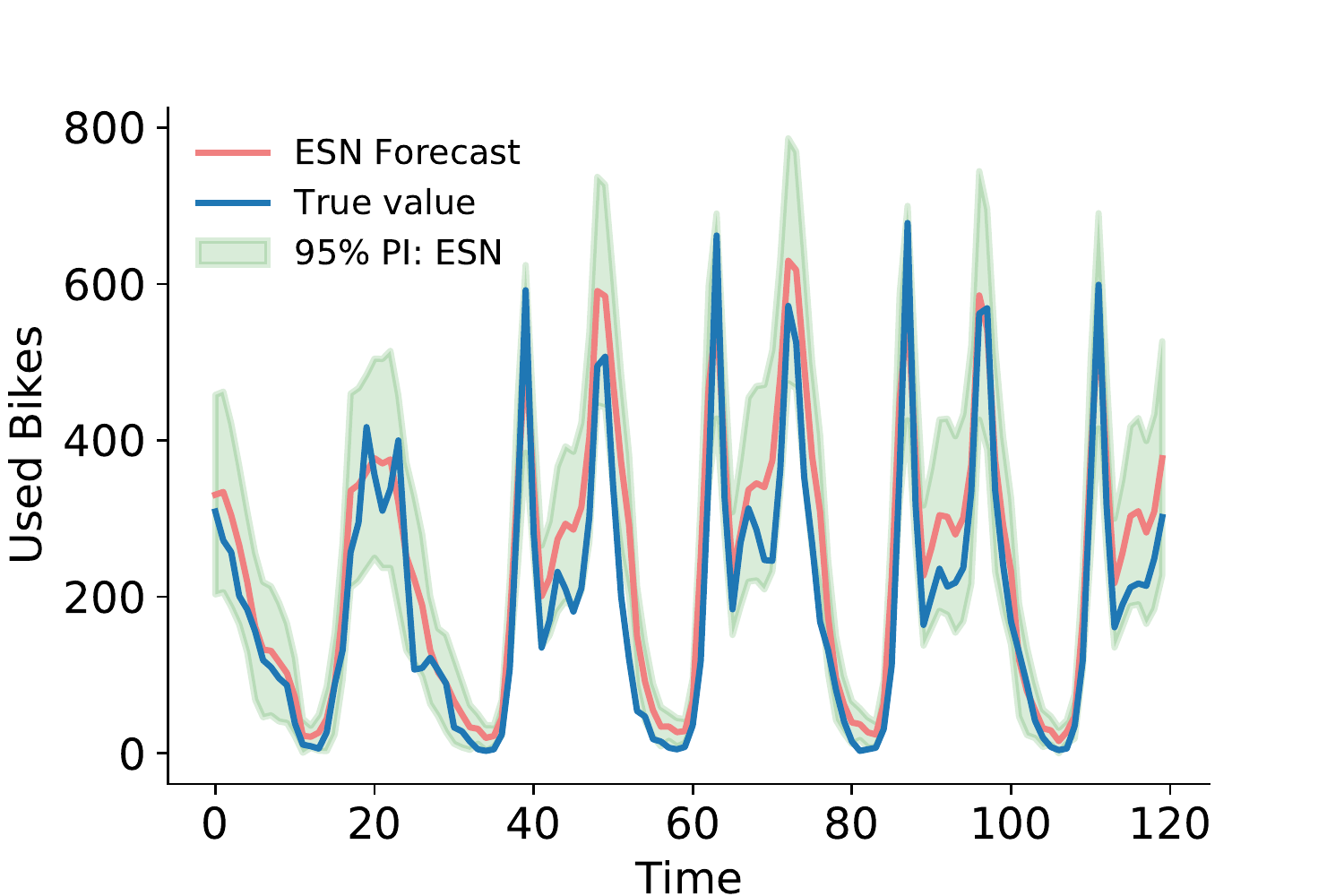}\caption[Daily RMSE]{Point Forecasts and Prediction Intervals for Bike Sharing Dataset.} 
	\label{fig:bikeresults} 
\end{figure}

Tables \ref{tab:results_02} to \ref{tab:results_06} detail the prediction interval results. We note that at least one deep learning algorithm is always in the top two performers. The interpretation is that deep learning algorithms outperform the traditional statistical methods, or at the very least, are competitive. This outperformance is also evidenced by the granular result for aggregated daily RMSE for HPC, as displayed in Figure~\ref{fig:dailyrmse}. Figure~\ref{fig:dailyrmse} is supplementary to Table~\ref{tab:results_04} as it gives a granular score at a daily level, which is consistent with the overall score detailed in the latter. Figure~\ref{fig:bikeresults} illustrates the point forecasts and prediction intervals for ESN (best performer for BSD) at the last $120$ hours of our forecast horizon. Our results are consistent with the findings of~\citet{Makri20a}, who noted that deep learning models produce better prediction intervals, albeit in the univariate case.

ESN is the best performing architecture for the BSD at the all prediction intervals considered. This result is consistent with the RMSE scores in Table \ref{tab:rmse}.

\subsection{Anomaly Detection}

\begin{figure*}[h!tbp]
    \centering
	\begin{subfigure}{0.325\textwidth}
		\centering 
		\includegraphics[width=\columnwidth]{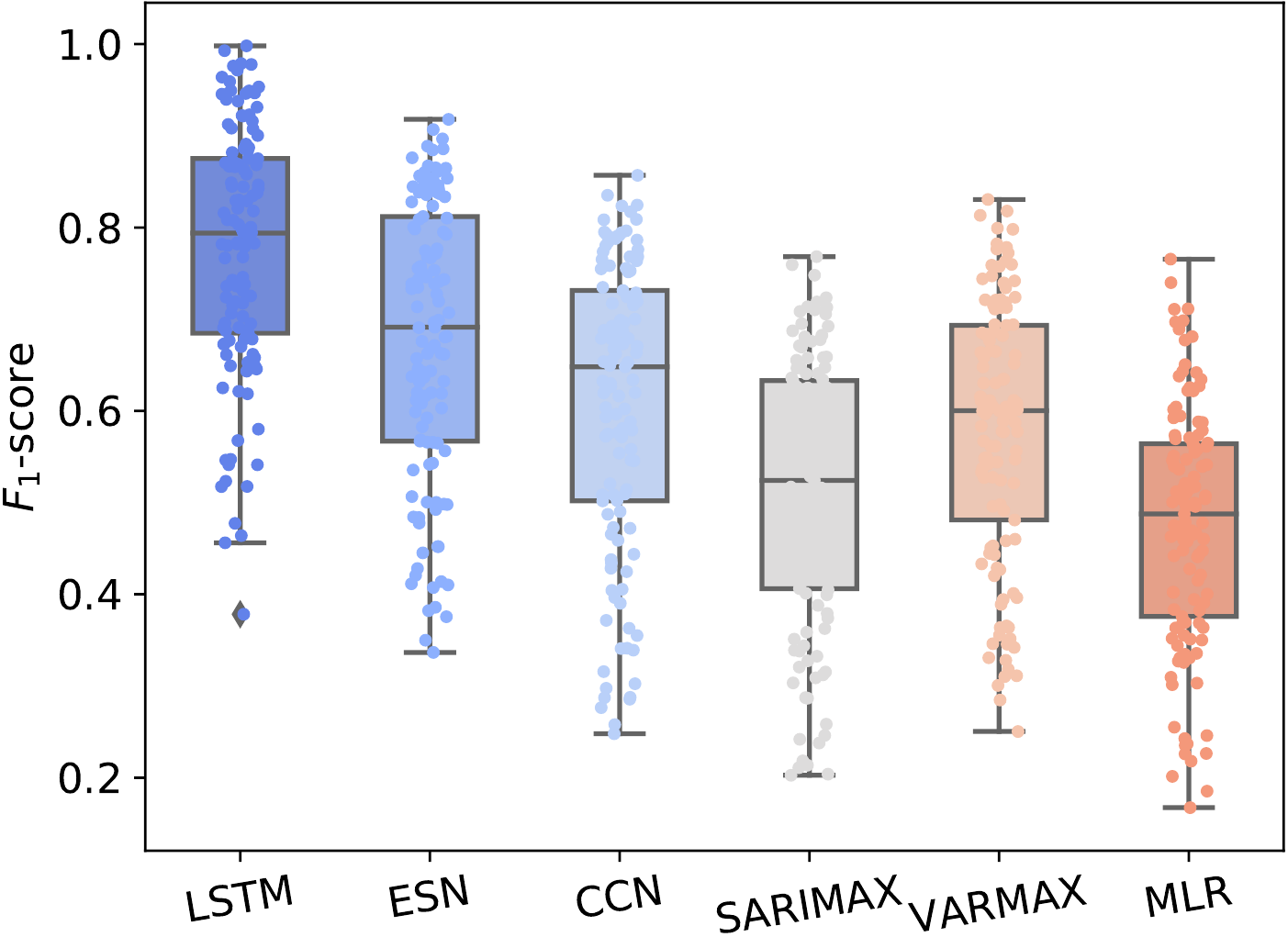}\caption{$F_1$-Score for AQD.} 
		\label{fig:f1a} 
	\end{subfigure} 
	\centering
	\begin{subfigure}{0.325\textwidth}
		\centering 
		\includegraphics[width=\columnwidth]{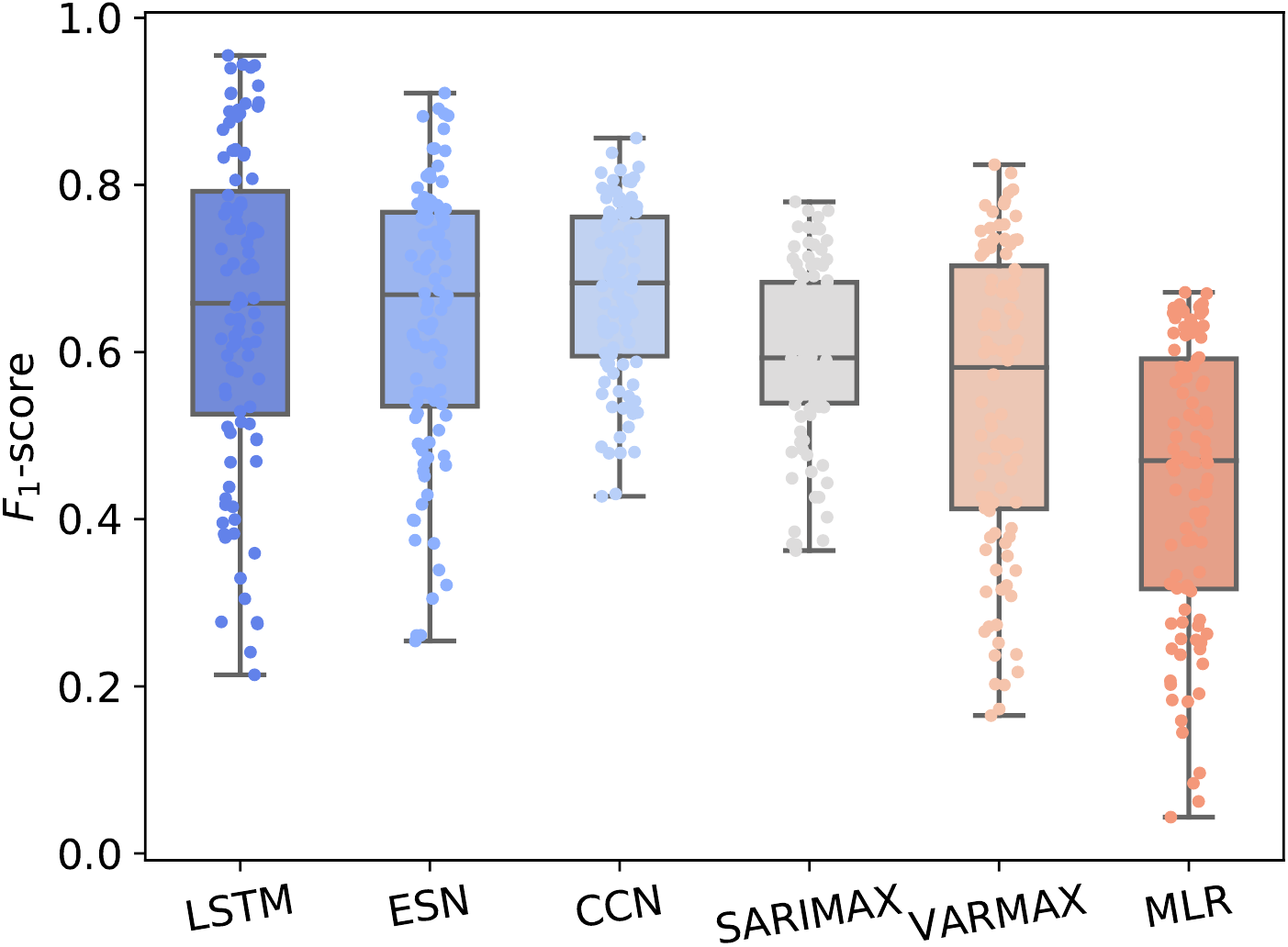}\caption{$F_1$-Score for BSD.} 
		\label{fig:f1b} 
	\end{subfigure} 
	\centering
	\begin{subfigure}{0.325\textwidth}
		\centering 
		\includegraphics[width=\columnwidth]{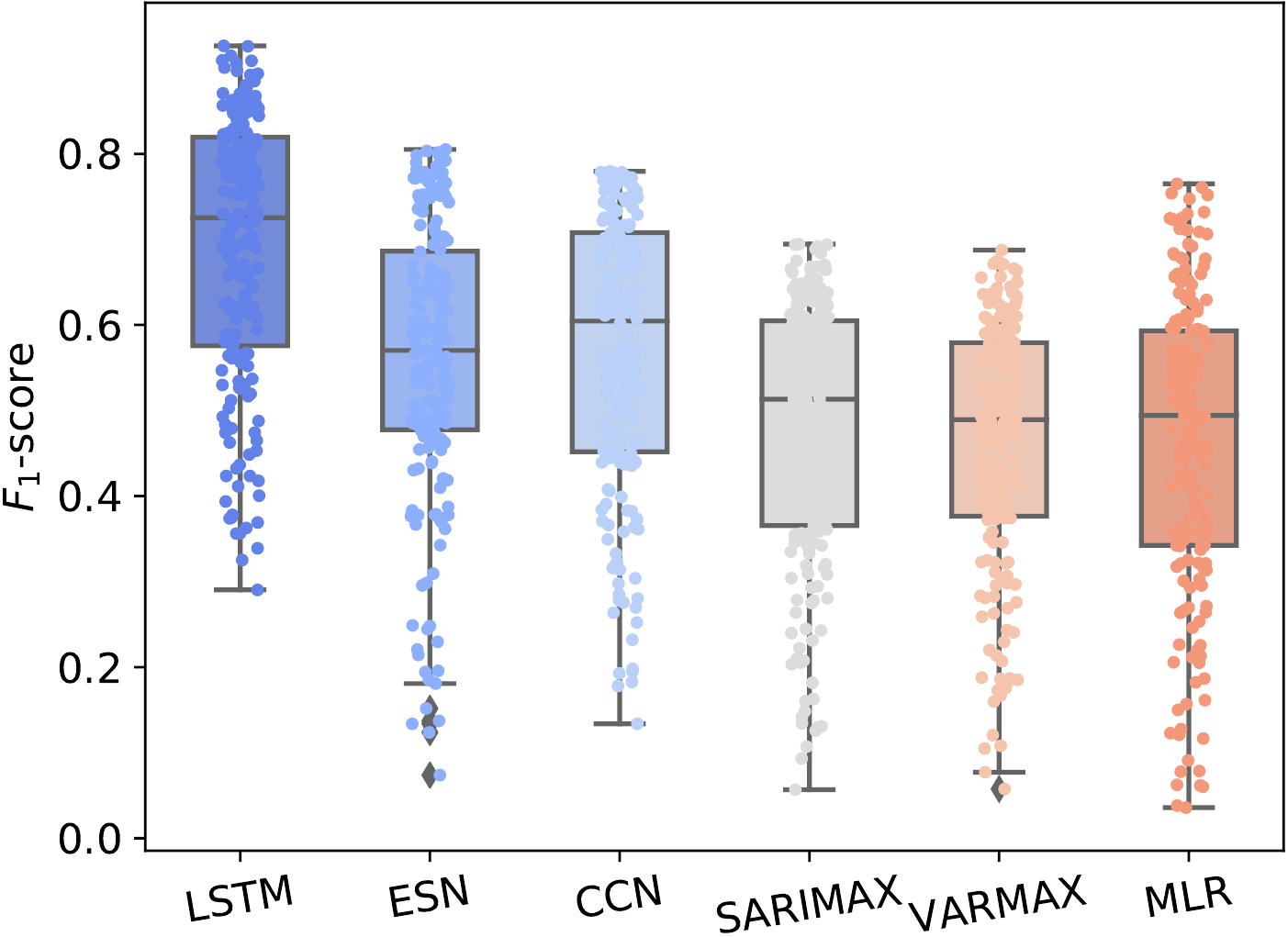}\caption[Daily RMSE]{$F_1$-Score for HPC.} 
		\label{fig:f1c}
	\end{subfigure}  \\
	\begin{subfigure}{0.325\textwidth}
		\centering 
		\includegraphics[width=\columnwidth]{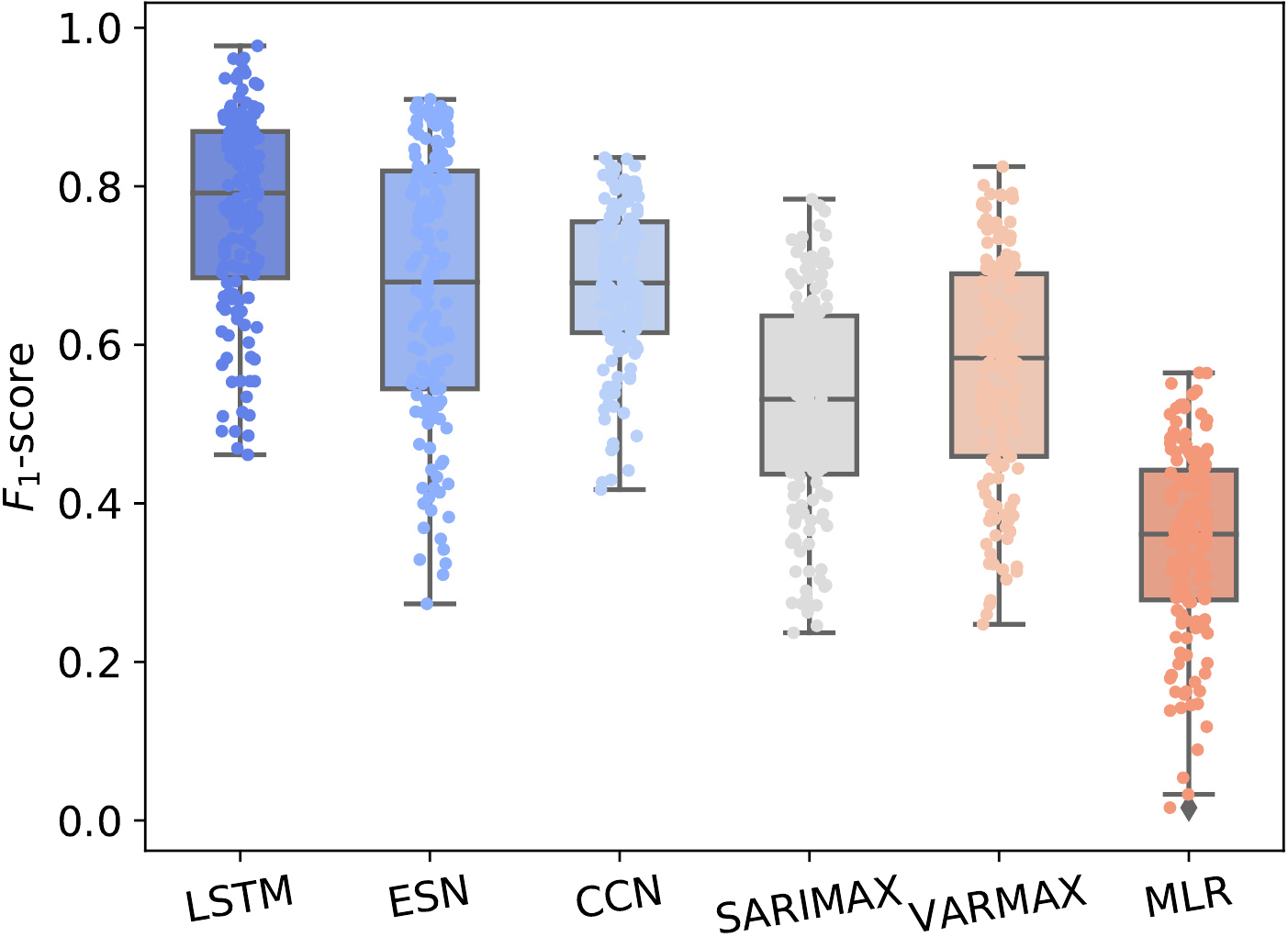}\caption[Daily RMSE]{$F_1$-Score for BDS.} 
		\label{fig:f1d}
	\end{subfigure} 
	\begin{subfigure}{0.325\textwidth}
		\includegraphics[width=\columnwidth]{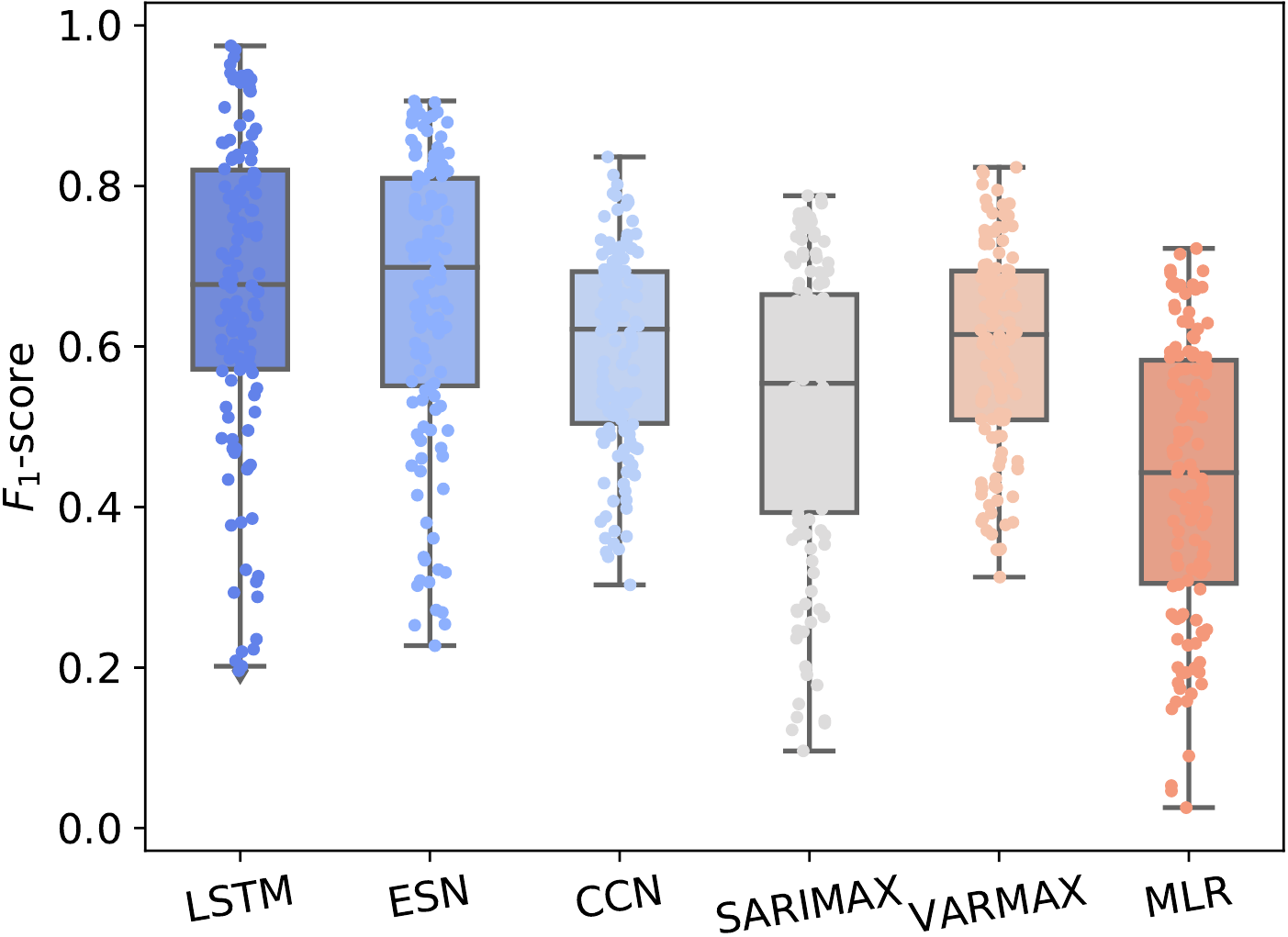}\caption[Daily RMSE]{$F_1$-Score for YWB.} 
		\label{fig:f1e} 
	\end{subfigure}
	\caption{$F_1$-Score distribution for each datasets and all the models.}
	\label{fig:f1}
\end{figure*}

\begin{figure*}[h!tbp]
	\centering
	\begin{subfigure}{0.325\columnwidth}
		\centering 
		\includegraphics[width=\columnwidth]{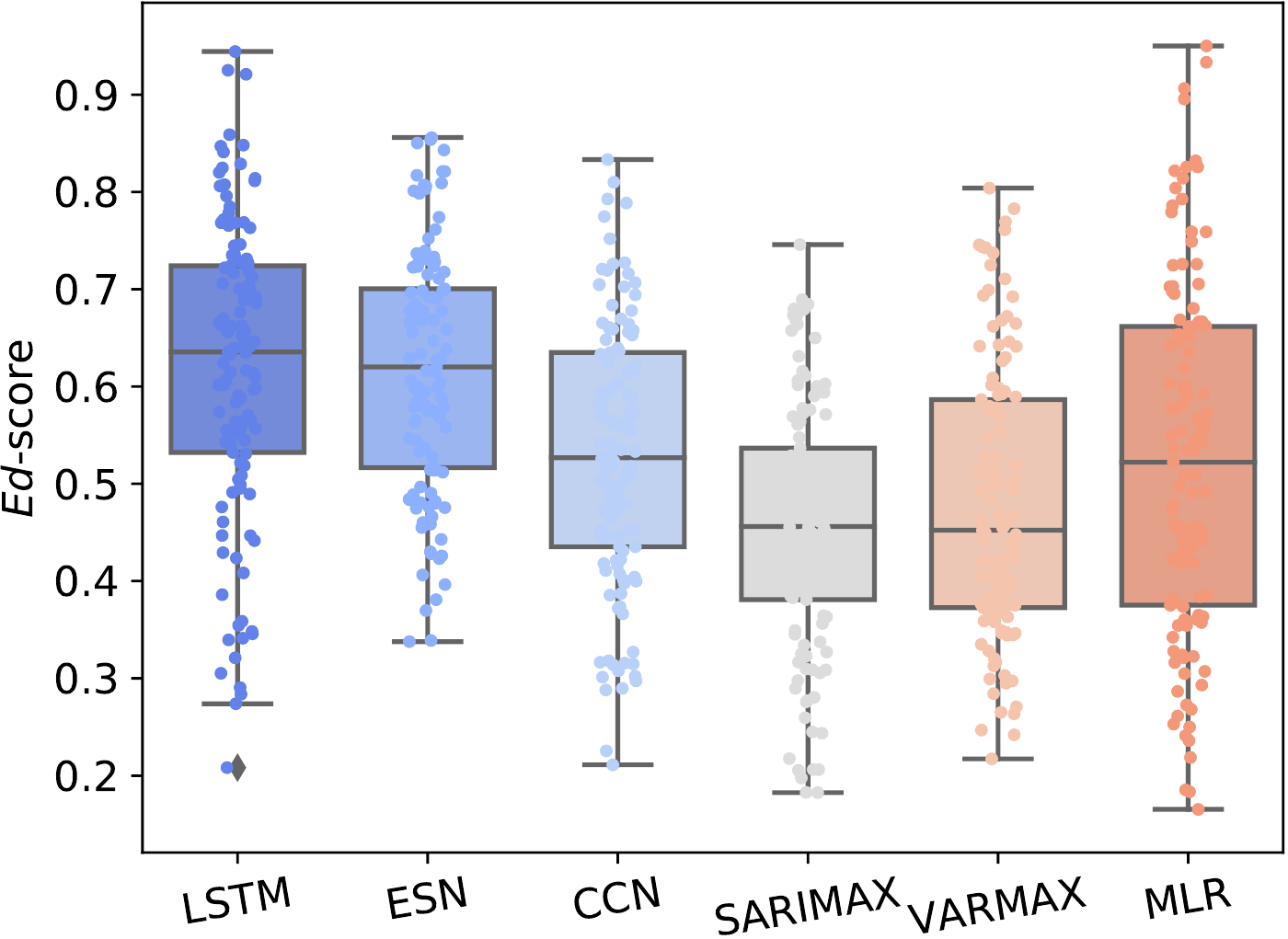}\caption{$Ed$-Score for AQD.} 
		\label{fig:Eda} 
	\end{subfigure} 
	\centering
	\begin{subfigure}{0.325\columnwidth}
		\centering 
		\includegraphics[width=\columnwidth]{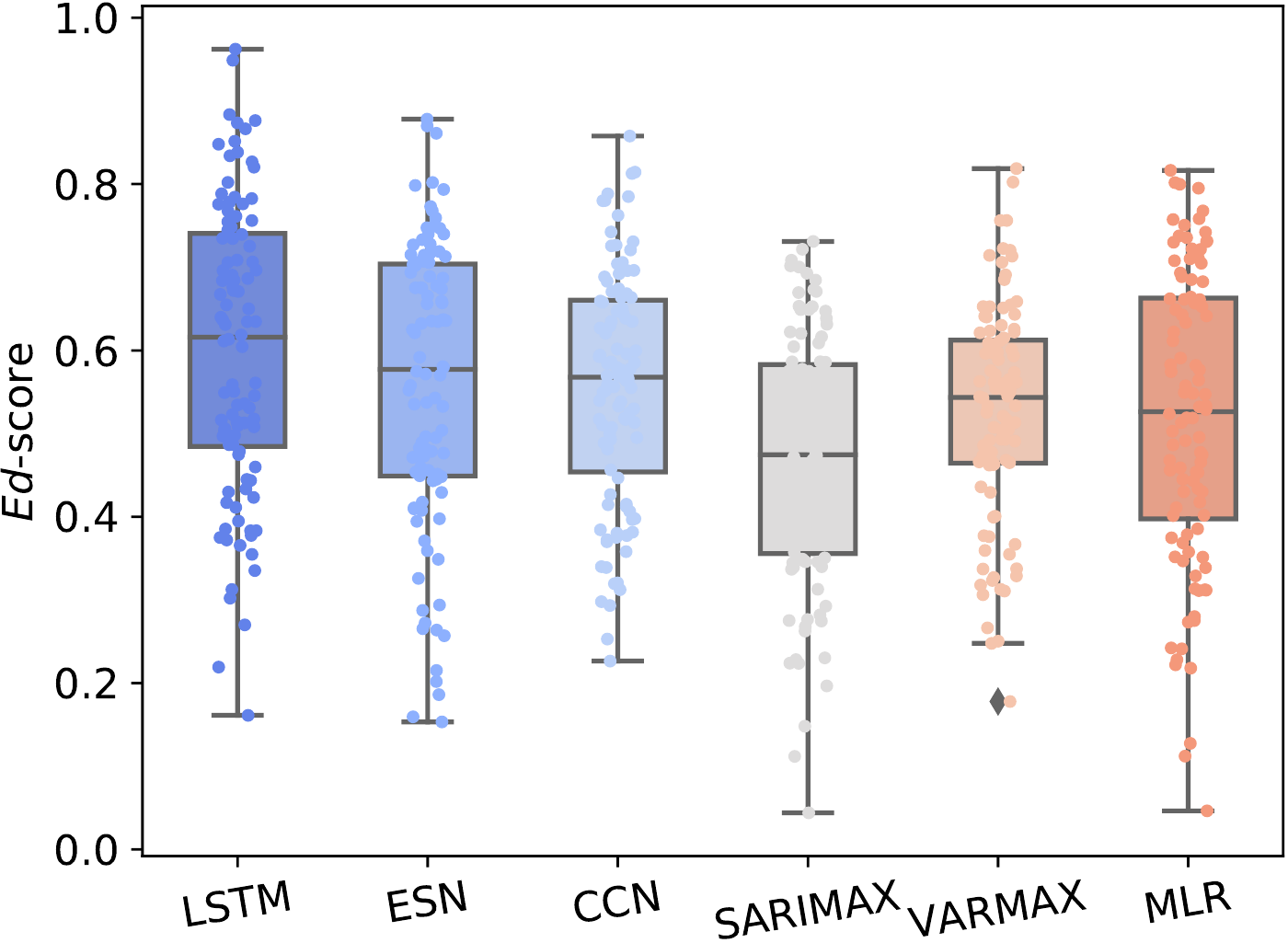}\caption{$Ed$-Score for BSD.} 
		\label{fig:Edb} 
	\end{subfigure} 
	\centering
	\begin{subfigure}{0.325\columnwidth}
		\centering 
		\includegraphics[width=\columnwidth]{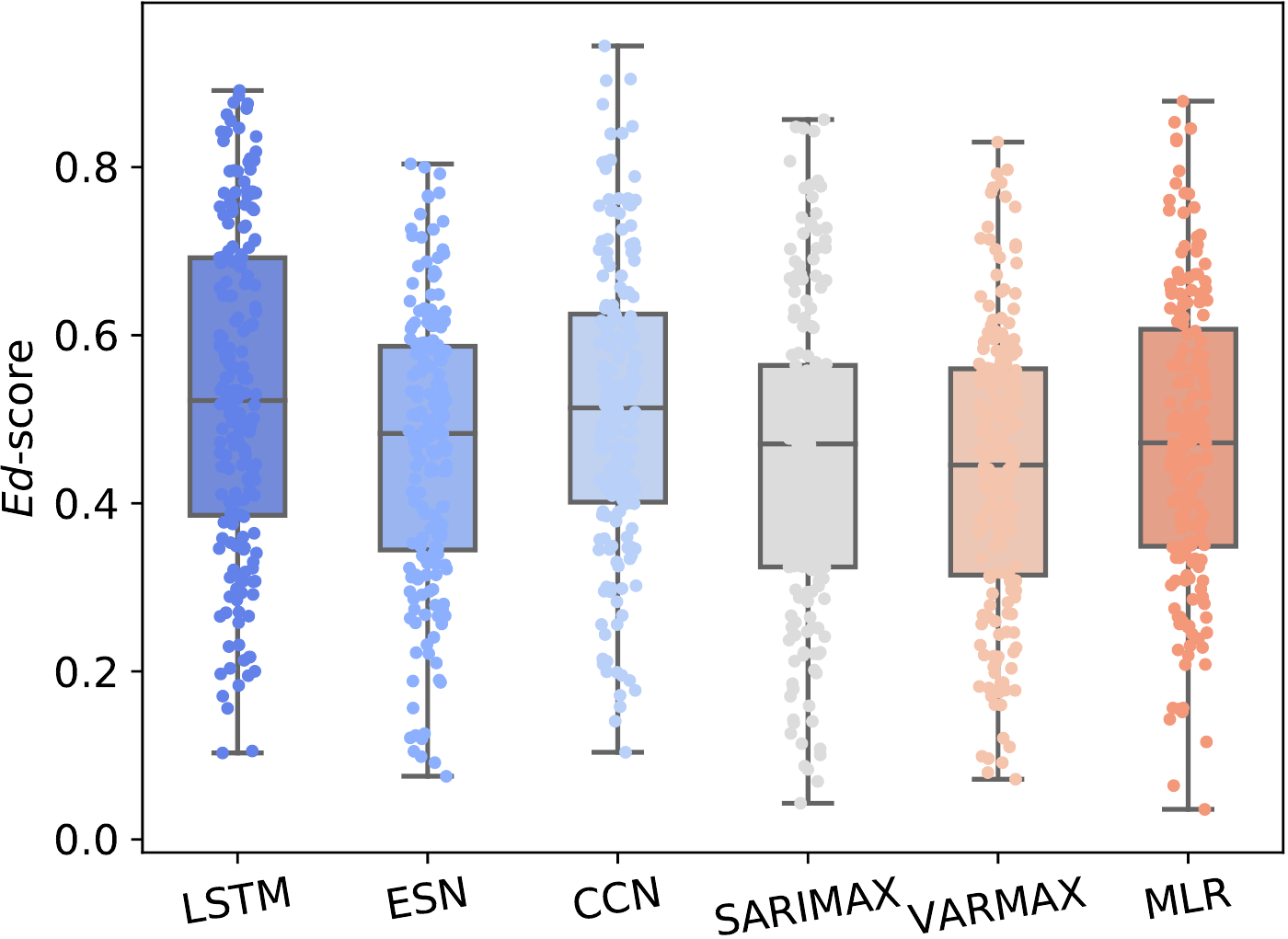}\caption[Daily RMSE]{$Ed$-Score for HPC.} 
		\label{fig:Edc}
	\end{subfigure} \\
	\begin{subfigure}{0.325\columnwidth}
		\centering 
		\includegraphics[width=\columnwidth]{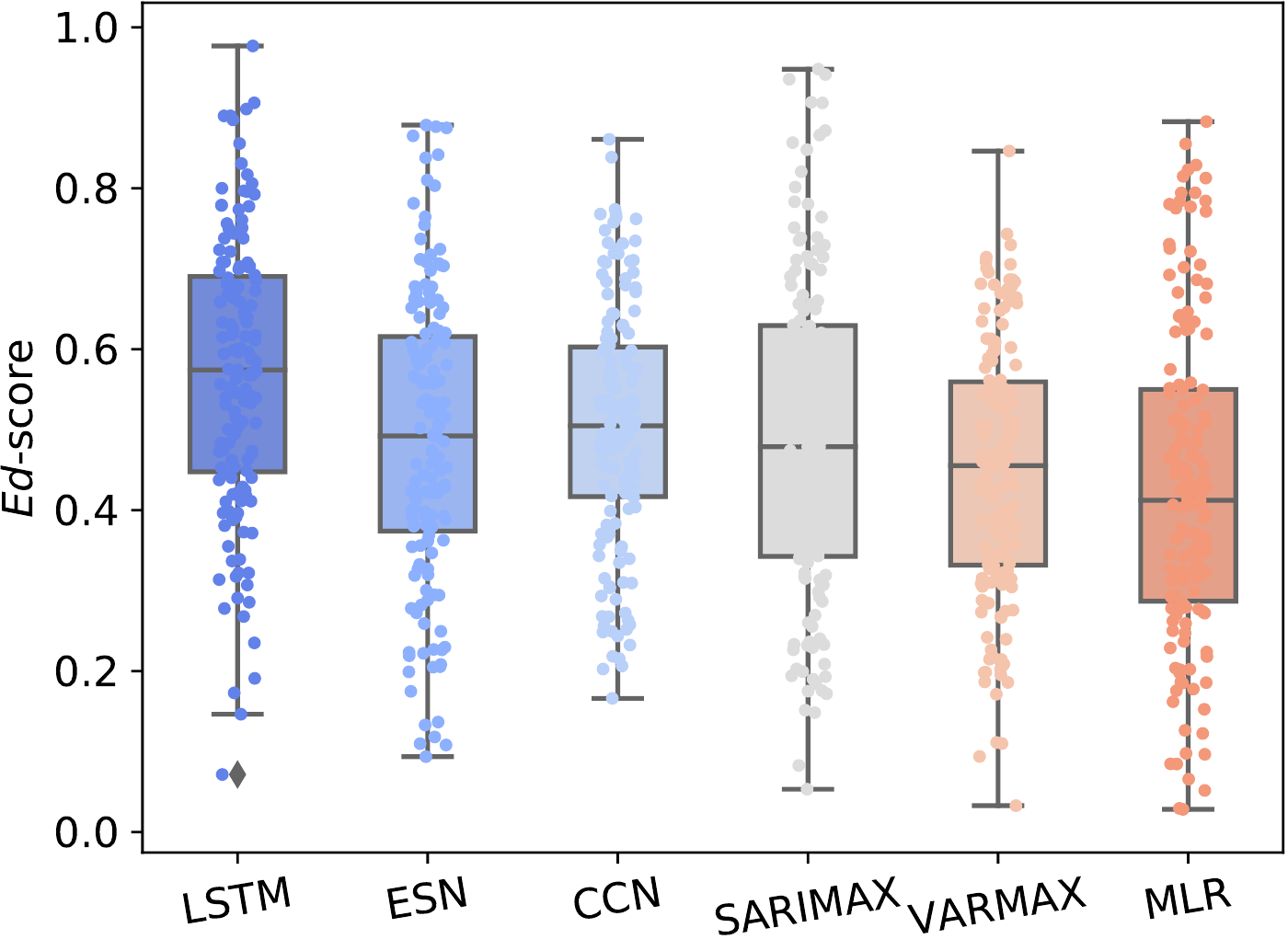}\caption[Daily RMSE]{$Ed$-Score for BDS.} 
		\label{fig:Edd}
	\end{subfigure} 
	\begin{subfigure}{0.325\columnwidth}
		\includegraphics[width=\columnwidth]{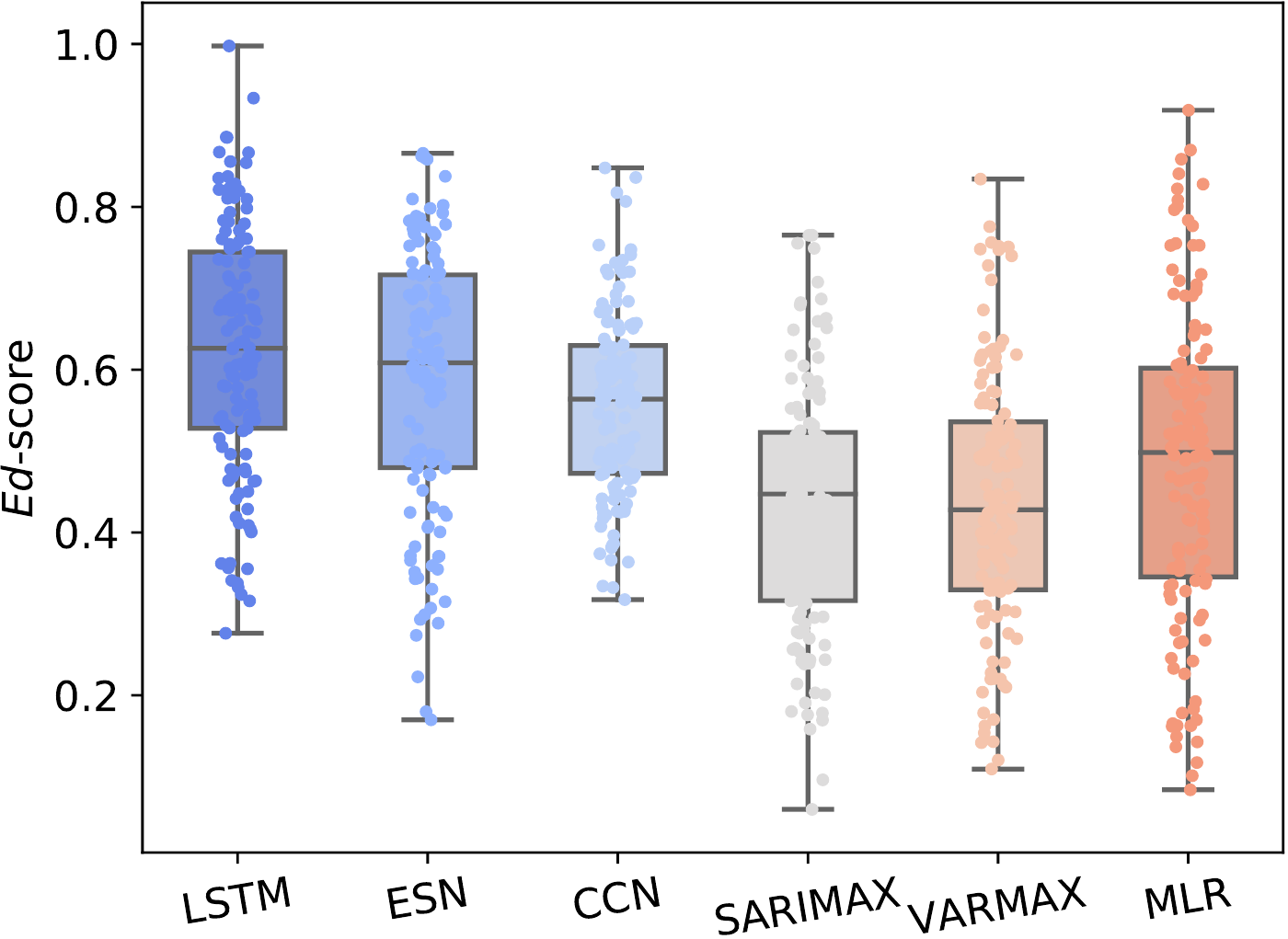}\caption[Daily RMSE]{$Ed$-Score for YWB.} 
		\label{fig:Ede} 
	\end{subfigure}
	\caption{$Ed$-Score distribution for each datasets and all the models.}
	\label{fig:Ed}
\end{figure*}

Figures~\ref{fig:f1} and~\ref{fig:Ed} detail the distributions for the $F_1$- and $Ed$-scores, respectively, for all the models applied to each dataset. We note the $F_1$-score is varied for each dataset, but the results indicate LSTM is the best performer, followed in every instance by either ESN or CCN. The worst performer is MLR across all datasets. Since the $F_1$-score is the harmonic average of Precision and Recall, where $1$ is the perfect score and $0$ is worst, we note that the deep learning models are better at learning the dynamic threshold.

Again, with the $Ed$-score, the statistical methods are outperformed by the deep learning models. This result follows as better dynamic threshold calibration leads to earlier detection of anomalies.

\section{Conclusion}

Our point of departure is that $(i)$ deep learning models have been shown to outperform traditional statistical methods at forecasting, and $(ii)$ deep learning has been shown to produce superior uncertainty quantification. In this paper, we $(i)$ produce point forecasts, $(ii)$ produce their associated prediction intervals using computational means where no built-in probabilistic constructs exist, and $(iii)$ use said prediction intervals to detect anomalies on different multivariate datasets.

With the methodology presented, we note  that better point forecast accuracy leads to the construction  of better prediction intervals, which in turn leads to the main result presented: better prediction intervals lead to more accurate and faster anomaly detection. The improvement is as a result of better parameterized models being capable of more accurate forecasts. These models in turn are used in the interval construction process and subsequently multivariate anomaly detection.

The goal was not to prove that deep learning is far superior to statistical methods in this regard, but to show that when proper methodology is followed, it can be competitive in the worst case. This in an important result, as it opens the way for the wide-scale adoption of deep learning in other applications.

The computational cost of the deep learning algorithms is not discussed in this paper and this could be an avenue for further exploration. The algorithmic time complexity is known for each model, with ESN being the fastest deep learning architecture. Even at that, it is still not as fast as any of the statistical methods. In terms of the deep learning architectures, future work can consider techniques for improving upon the cost of computation, so that even in this aspect they become competitive to their statistical counterparts.

Finding techniques to address this last issue in particular, will enable researchers to study multivariate time series in more detail, and at a grander scale. There exists a need for conducting our kind of analysis on multivariate time series at the scale of, say, the M4 competition \cite{Makri20a}.

\section*{Conflict of Interest}

The authors declare that they have no conflict of interest.

\bibliography{references_03.bib}
\bibliographystyle{plainnat} 

\end{document}